\PassOptionsToPackage{table}{xcolor}
\documentclass{fairmeta}

\usepackage[utf8]{inputenc}
\usepackage{float}
\usepackage{tabularx}
\usepackage{array}
\usepackage{wrapfig}
\usepackage{needspace}
\usepackage{adjustbox}
\usepackage{titletoc}
\usepackage{amssymb}

\usepackage{amsmath,amsfonts,bm}

\def\eqref#1{equation~\ref{#1}}

\def\1{\bm{1}}

\DeclareMathAlphabet{\mathsfit}{\encodingdefault}{\sfdefault}{m}{sl}
\SetMathAlphabet{\mathsfit}{bold}{\encodingdefault}{\sfdefault}{bx}{n}

\newcolumntype{Y}{>{\centering\arraybackslash}X}

\title{MeSD: Multi-Evidence Self-Distillation for VideoLLM}
\renewcommand\authorformat[2][]{\mbox{\sffamily\bfseries #2$^{#1}$}}
\renewcommand\affiliationformat[2][]{\mbox{\small $^{#1}$#2}}
\newcommand\authornewrow[2][]{\addtolist[#1]{#2}{\authorlist}{\authorformat}{,\protect\newline}}
\newcommand\affiliationnewrow[2][]{\addtolist[#1]{#2}{\affiliationlist}{\affiliationformat}{,\protect\newline}}
\author[1,2,\dagger]{Weijie Zhu}      
\author[2,\dagger]{Han Fang}        
\author[1]{Hanyu Fu}                
\author[3]{Yuzhe Zhang}             
\author[2]{Xin Wei}                 
\authornewrow[4]{Zhaoyan Pan}       
\author[5]{Feiran Liu}              
\author[6]{Xunjie Jin}              
\author[2]{Hongbo Sun}              
\author[2]{Zhiyu Lin}               
\author[2]{Tianyi Gao}              
\authornewrow[2]{Tianyi Ding}       
\author[2]{Ye Yuan}                 
\author[2]{Zhongjiang He}           
\author[2,*]{Hao Sun}               
\author[1,*]{Zhiheng Wu}            

\affiliation[1]{UCAS}
\affiliation[2]{Xingchen AGI Lab, China Telecom Artificial Intelligence Technology (Beijing) Co., Ltd}

\affiliationnewrow[3]{PKU}
\affiliation[4]{ZJU}
\affiliation[5]{BJTU}
\affiliation[6]{BUAA}

\contribution[\dagger]{Equal contribution}
\contribution[*]{Corresponding authors.}

\metadata[Project page]{\href{https://whiteventral.github.io/MeSD/}{%
  \raisebox{-2pt}{\includegraphics[height=10pt]{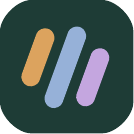}}\hspace{0.45em}%
  \nolinkurl{https://whiteventral.github.io/MeSD/}}}
\patchcmd{\mymaketitle}{\vskip 0.5cm}{\vskip 0.12cm}{}{%
  \ClassError{fairmeta}{Cannot set project-link spacing}{Check the title-box definition.}}
\patchcmd{\mymaketitle}{\setlength{\parskip}{0.5cm}}{\setlength{\parskip}{0.2cm}}{}{%
  \ClassError{fairmeta}{Cannot set title-box paragraph spacing}{Check the title-box definition.}}
\patchcmd{\mymaketitle}{\tcbset{bottom=0.35cm}}{\tcbset{bottom=0.15cm}}{}{%
  \ClassError{fairmeta}{Cannot set title-box bottom padding}{Check the title-box definition.}}
\hypersetup{
  pdftitle={MeSD: Multi-Evidence Self-Distillation for VideoLLM},
  pdfauthor={Weijie Zhu, Han Fang, Hanyu Fu, Yuzhe Zhang, Xin Wei, Zhaoyan Pan, Feiran Liu, Xunjie Jin, Hongbo Sun, Zhiyu Lin, Tianyi Gao, Tianyi Ding, Ye Yuan, Zhongjiang He, Hao Sun, Zhiheng Wu}
}

\begin{document}
\begingroup
\setlength{\parskip}{4pt plus 1pt}
\setlength{\textfloatsep}{12pt plus 2pt minus 2pt}
\setlength{\floatsep}{10pt plus 2pt minus 2pt}
\setlength{\intextsep}{10pt plus 2pt minus 2pt}
\setlength{\abovedisplayskip}{7pt plus 2pt minus 2pt}
\setlength{\belowdisplayskip}{7pt plus 2pt minus 2pt}
\setlength{\abovedisplayshortskip}{3pt plus 1pt}
\setlength{\belowdisplayshortskip}{5pt plus 1pt minus 1pt}
\titlespacing*{\section}{0pt}{12pt plus 2pt minus 2pt}{6pt plus 1pt}
\titlespacing*{\subsection}{0pt}{9pt plus 2pt minus 2pt}{4pt plus 1pt}
\captionsetup{skip=4pt}
\abstract{

While reinforcement learning with verifiable rewards provides reliable outcome supervision for VideoLLMs, sequence-level rewards offer limited token-level guidance. On-policy self-distillation addresses this limitation by conditioning a self-teacher on privileged information to provide dense token-level supervision. 
However, aggregating heterogeneous evidence within a single teacher context obscures cross-evidence agreement and conflict.
A further challenge lies in determining whether teacher guidance should refine reward-based updates or provide corrective supervision for failed trajectories.
To address these issues, we propose MeSD, a multi-evidence self-distillation framework for VideoLLMs. MeSD constructs three evidence-conditioned teachers with shared parameters, using the ground-truth answer as a common semantic context while separately incorporating temporal and spatial evidence. 
Given the same  student-generated prefixes, MeSD evaluates evidence-specific preferences relative to the Answer Teacher and fuses teacher-common preferences with gated teacher-specific residuals. Furthermore, MeSD introduces Verification-Guided Optimization to classify trajectories as Success, Failure, or Indeterminate. For Success and Indeterminate trajectories, MeSD refines token-level advantage magnitudes while preserving reward-derived signs. 
For verified failure trajectories that contain the required evidence, MeSD applies failure-conditioned distillation, using reverse-KL correction toward the fused distribution. 
Experiments on multiple video benchmarks demonstrate consistent gains over reinforcement learning and self-distillation baselines.

}
\maketitle

\section{Introduction}
\label{sec:introduction}

Video Large Language Models (VideoLLMs) have made significant progress in video question answering and temporal understanding~\citep{bai2025qwen25vl,zhang2025videollama3,
ren2024timechat,wang2025timer1}. However, robust reasoning over complex videos remains challenging~\citep{meng2025openo3video,lin2026visd}. Such reasoning requires models to temporally identify key events, spatially localize relevant objects, and integrate cross-frame observations to derive the final answer~\citep{meng2025openo3video}.
Final-answer rewards cannot reliably indicate whether intermediate steps correctly use temporal and spatial evidence, making token-level supervision critical~\citep{lin2026visd}.

Reinforcement learning with verifiable rewards (RLVR) provides reliable outcome supervision~\citep{guo2025deepseek}, but sequence-level rewards offer limited token-level guidance~\citep{guo2026segment}.
On-policy distillation (OPD)~\citep{ma2026mopd} provides token-level supervision on student-generated trajectories, but typically depends on a stronger teacher.
On-policy self-distillation (OPSD)~\citep{zhao2026opsd} avoids this dependence by using the student model conditioned on privileged information as its teacher.
However, matching the privileged teacher's distribution may lead to privileged information leakage and training instability.
RLSD~\citep{yang2026rlsd} therefore uses rewards to determine the signs of token-level advantages and teacher signals to refine their magnitudes.

\begin{figure}[t]
    \centering
    \includegraphics[width=\textwidth]{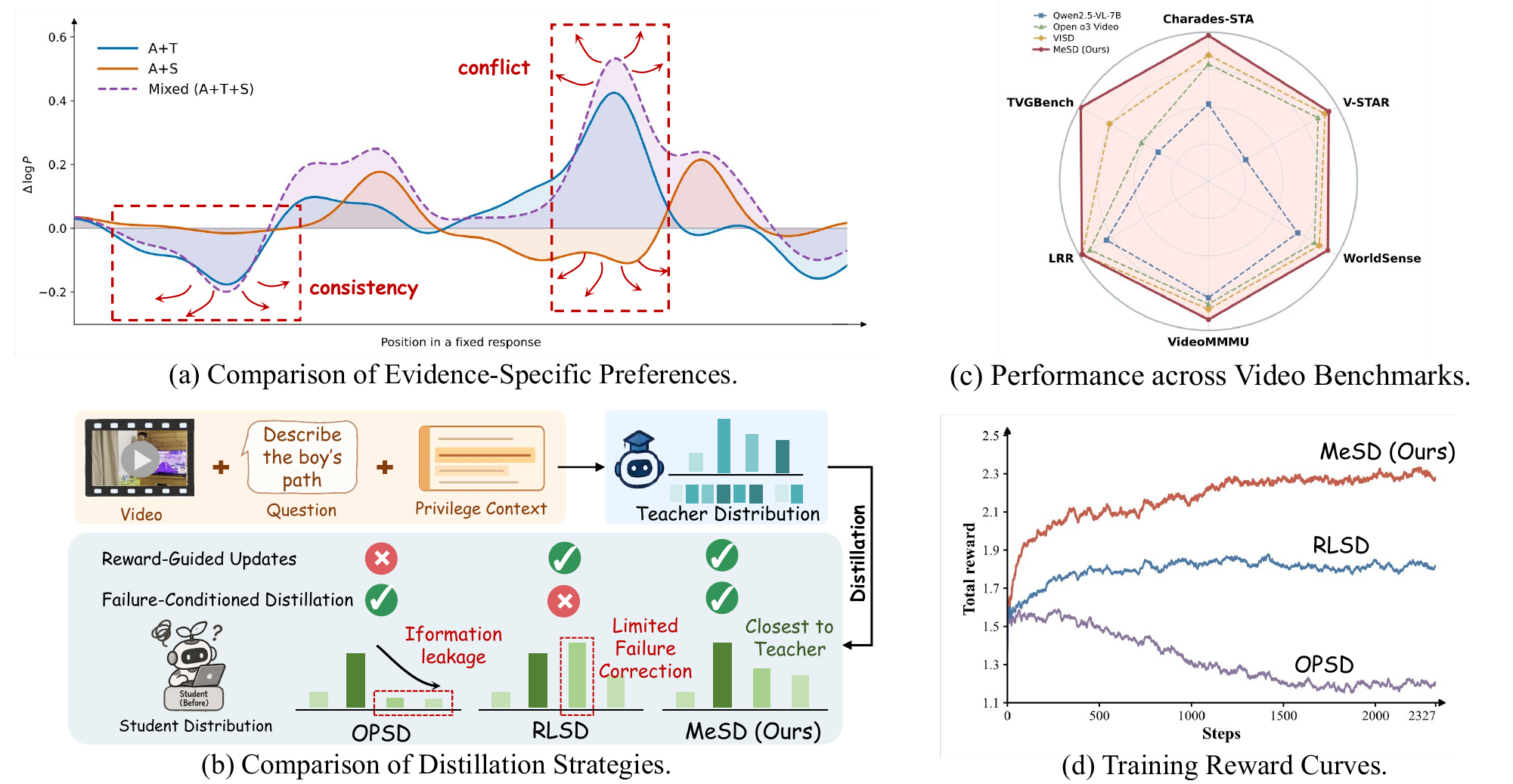}
    \caption{(a) Temporal and spatial evidence yield consistent or conflicting token preferences. (b) MeSD combines reward-anchored credit refinement with corrective distillation on verified failures. (c) MeSD leads on selected metrics across six benchmarks. (d) Comparison of training rewards.}
    \label{intro}
\end{figure}

However, applying self-distillation to video reasoning exposes two challenges.
First, heterogeneous privileged evidence can induce conflicting token preferences along the same reasoning trajectory, as illustrated in Figure~\ref{intro}(a). 
Combining these sources within a single teacher context obscures their individual contributions, as well as any agreement or conflict between them. 
Second, existing methods like RLSD use teacher signals to adjust token-level advantage magnitudes while preserving reward-derived signs across both successful and failed trajectories. 
As shown in Figure~\ref{intro}(b), this merely reweights updates to sampled tokens without directly aligning the student distribution with the teacher distribution, thereby offering limited corrective guidance for failed trajectories. 
These challenges raise two questions: how to model heterogeneous privileged evidence and when teacher guidance should refine reward-based updates or correct failed trajectories.

To address these limitations, we propose \textbf{MeSD}, a multi-evidence self-distillation framework for VideoLLMs.
MeSD integrates evidence-specific preferences and guides optimization based on trajectory verification outcomes and evidence availability.
It comprises two components: Multi-Evidence Self-Teacher and Verification-Guided Optimization.
The Multi-Evidence Self-Teacher employs three evidence-conditioned teachers with shared parameters: the Answer, Answer-Temporal, and Answer-Spatial Teachers.
The ground-truth answer provides a common semantic context, while temporal and spatial evidence are introduced separately.
Using the Answer Teacher distribution as a semantic anchor, MeSD captures the token preference shifts induced by additional temporal or spatial evidence.
It then combines teacher-common preferences with gated teacher-specific residuals to yield a fused teacher distribution.
The fusion selectively retains teacher-specific support while reducing residual contributions when evidence-specific preferences conflict.

Verification-Guided Optimization uses a task-specific verifier to classify trajectories as Success, Failure, or Indeterminate.
For Success and Indeterminate trajectories, MeSD retains token-level advantage refinement, adjusting advantage magnitudes while preserving their reward-derived signs.
MeSD applies failure-conditioned distillation only to verified failures with the required temporal or spatial evidence, using a reverse-KL objective targeting the fused teacher distribution.
This objective uses teacher preferences over candidate tokens to provide corrective guidance beyond sampled-token reweighting. Overall, MeSD improves performance across video benchmarks and achieves sustained reward gains during training, as shown in Figure~\ref{intro}(c,d).
Our main contributions are:

\begin{itemize}
\setlength{\itemsep}{2pt}
\setlength{\parsep}{0pt}
\item We introduce MeSD, to our knowledge the first multi-teacher on-policy self-distillation framework for VideoLLMs.
Its Multi-Evidence Self-Teacher uses the Answer Teacher as a reference to fuse teacher-common preferences with gated teacher-specific residuals.

\item We propose Verification-Guided Optimization, which refines token-level advantage magnitudes while preserving reward-derived signs for Success and Indeterminate trajectories. It applies failure-conditioned distillation only to verified failures with the required evidence, using a reverse-KL objective targeting the fused teacher distribution.

\item Extensive experiments across multiple video reasoning benchmarks demonstrate that MeSD consistently outperforms RLVR, OPSD, RLSD, and video post-training methods.
\end{itemize}

\clearpage
\section{Related Work}
\label{sec:related-work}

\subsection{Post-Training for VideoLLMs}
\label{sec:video_post_training}

RLVR improves video understanding and reasoning through verifiable outcome feedback~\citep{guo2025deepseek,Wu_2026_CVPR,wang2025videorft}.
Video-R1~\cite{feng2025videor1} and Time-R1~\cite{wang2025timer1} target temporal reasoning and grounding, respectively.
GKD~\cite{agarwal2024gkd} and Video-OPD~\cite{li2026videoopd} supervise student-generated trajectories using teacher distributions to mitigate training--inference mismatch.
OPSD~\cite{zhao2026opsd} and SDPO~\cite{hubotter2026sdpo} condition self-teachers on privileged solutions and environment feedback, respectively.
Visual variants use regional crops (Vision-OPD~\cite{yuan2026visionopd}), annotated evidence frames (Video-OPSD~\cite{wang2026videoopsd}), or denser temporal sampling (S$^3$T~\cite{venkatraman2026s3t}).
Open-o3 Video~\cite{meng2025openo3video} jointly supervises answers and spatiotemporal evidence through reasoning traces containing timestamps, objects, and bounding boxes. Its rewards jointly evaluate answer correctness and spatiotemporal grounding quality.
Following this paradigm, MeSD provides token-level guidance for student-generated trajectories.

\subsection{Multi-Teacher Distillation and Optimization}
\label{sec:multi_view_distillation}

MOPD~\cite{ma2026mopd} combines independently trained specialists, while MT-SDPO~\cite{he2026mtsdpo} conditions an EMA self-teacher on verified teacher feedback and AVSD~\cite{nguyen2026avsd} aggregates privileged views.
For optimization, VG-OPD~\cite{xu2026vgopd} uses verification to select eligible experts and localize token-level supervision.
RLSD~\cite{yang2026rlsd} and VISD~\cite{lin2026visd} refine token-level credit while preserving reward-derived advantage signs, whereas H$^2$SD~\cite{cai2026h2sd} uses outcome-dependent asymmetric supervision.
MeSD measures temporal and spatial preference shifts relative to an answer-conditioned anchor, then fuses teacher-common preferences with gated teacher-specific residuals to account for cross-evidence agreement and conflict.
It uses verification outcomes and evidence availability to choose between reward-anchored refinement and reverse-KL correction, reserving the latter for eligible verified failures.
\section{Method}
\subsection{Overview}

We propose MeSD, a multi-evidence self-distillation framework designed for VideoLLMs. The core challenge is to integrate heterogeneous privileged evidence into token-level supervision and determine how this supervision should guide trajectories with different verification outcomes.
As illustrated in Figure~\ref{main}, MeSD constructs Evidence-Conditioned Teachers from answer, temporal, and spatial annotations, with shared EMA parameters. These teachers evaluate reasoning trajectories sampled by the
student from video-question pairs.
Teacher Distribution Fusion compares their predictions at the same student-generated prefixes and aggregates evidence-specific preferences into a fused teacher distribution.
Verification-Guided Optimization uses a task-specific verifier to classify each trajectory as Success, Failure, or Indeterminate. For Success and Indeterminate trajectories, token-level advantage refinement adjusts advantage magnitudes while preserving their reward-derived signs. 
For verified failures with the required evidence, failure-conditioned distillation replaces the GRPO policy loss with a reverse-KL objective targeting the fused teacher distribution.

\begin{figure}[tp]
    \centering
    \includegraphics[width=\textwidth]{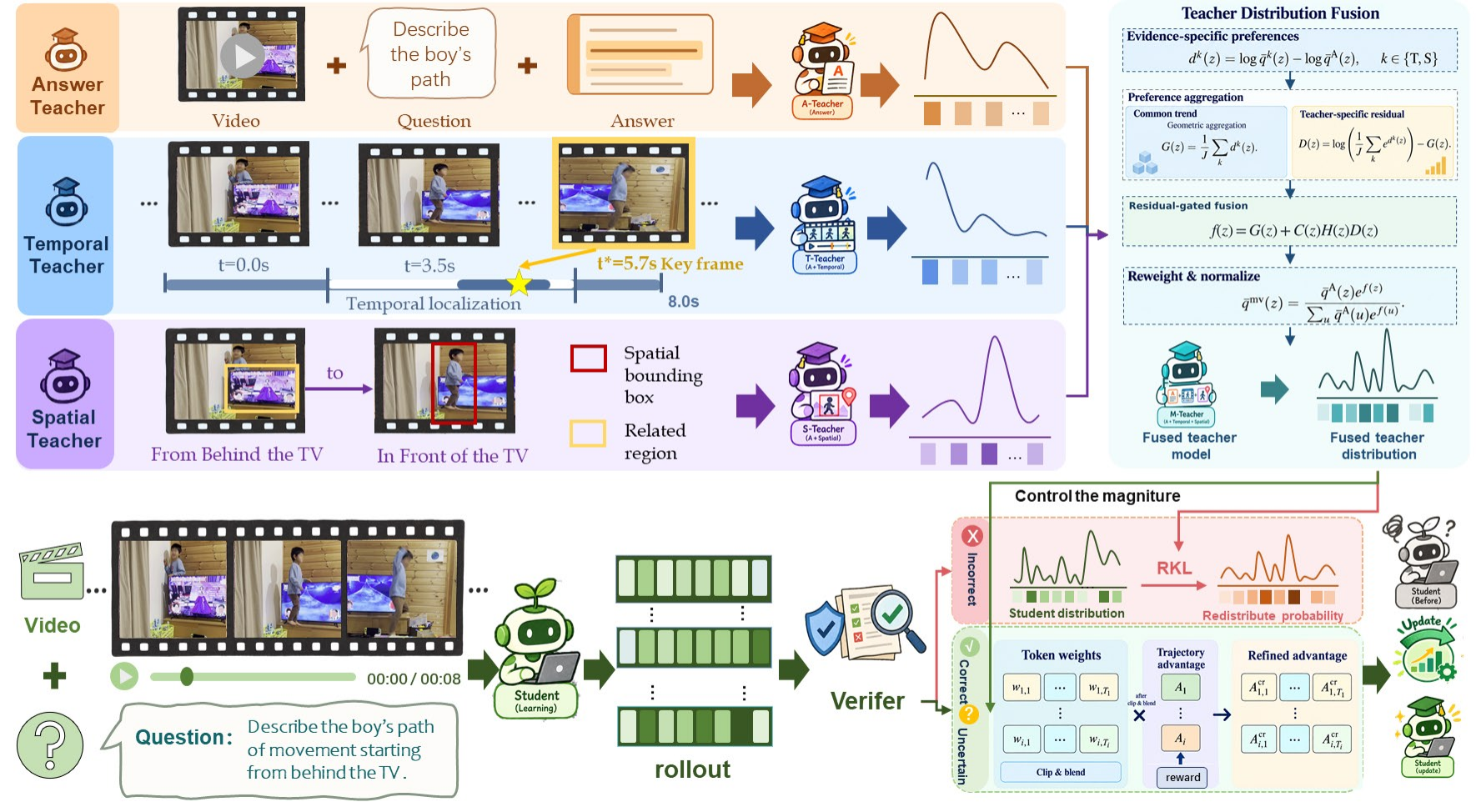}
    \caption{Overview of MeSD. Teachers sharing EMA parameters use
answer, temporal, and spatial annotations as privileged information
to evaluate student-generated trajectories.
Using the Answer Teacher as an anchor, teacher distribution fusion
aggregates evidence-specific preferences through a gated
consensus--residual decomposition.
Trajectory verification determines how the student learns.
For failed trajectories, the student
matches the fused teacher distribution through reverse-KL distillation.
For other trajectories, teacher guidance adjusts the strength
of token-level updates, and rewards determine whether tokens
are reinforced or suppressed.}
    \label{main}
\end{figure}

\subsection{Multi-Evidence Self-Teacher}

As illustrated in Figure~\ref{intro}, temporal and spatial evidence can drive consistent or conflicting token preferences along a single reasoning trajectory.
Integrating these evidence sources within a single teacher context obscures their relationship and makes their individual contributions
hard to distinguish.
To address this, we construct distinct evidence-conditioned teachers and fuse their predictions through teacher-common preferences and gated teacher-specific residuals.

\noindent\textbf{Evidence-Conditioned Teachers.}
Given a video-question pair $x_i$ and a student-generated prefix $y_{i,<t}$, we construct three evidence-conditioned teachers with shared parameters $\bar\theta$.
The ground-truth answer provides a common semantic context, with temporal and spatial evidence added separately:
\begin{equation}
\begin{gathered}
\bar q_{i,t}^{k}
=
\pi_{\bar\theta}
\bigl(\cdot\mid x_i,E_i^k,y_{i,<t}\bigr),
\quad k\in\{\mathrm A,\mathrm T,\mathrm S\},\\
E_i^{\mathrm A}=a_i,\quad
E_i^{\mathrm T}=(a_i,\tau_i),\quad
E_i^{\mathrm S}=(a_i,\mathcal B_i),
\end{gathered}
\end{equation}
where $\mathrm A$, $\mathrm T$, and $\mathrm S$ denote the Answer, Answer-Temporal, and Answer-Spatial Teachers, respectively.
The component $a_i$ contains only the semantic ground-truth answer, excluding temporal intervals, keyframe timestamps, and bounding box coordinates.
The temporal evidence $\tau_i$ provides annotated time intervals or keyframe timestamps.
For the spatial evidence, we format the ground-truth answer, object names, and transformed bounding boxes as privileged information.
The teachers compute predictions for the existing student trajectories without generating additional rollouts, and these predictions are detached to prevent gradient backpropagation.

\noindent\textbf{Teacher Distribution Fusion.}
Given these teacher distributions, we use the Answer Teacher as the anchor teacher and measure
evidence-specific preferences relative to its predictions.
We then combine \textit{teacher-common preferences} with a \textit{gated teacher-specific residual} to capture consistency and conflict across evidence sources.
For each candidate token $z$, we define the preference change induced by temporal or spatial evidence as:
\begin{equation}
d^k(z)=\log\bar q^k(z)-\log\bar q^{\mathrm A}(z),
\qquad k\in\{\mathrm T,\mathrm S\},
\end{equation}
where $\bar q^k$ denotes $\bar q_{i,t}^k$ with the sample and token indices omitted.
A positive $d^k(z)$ indicates that the additional evidence increases the support for token $z$ relative to the Anchor Teacher, while a negative value indicates a decrease in support.

The geometric mean emphasizes tokens supported across all non-anchor teachers, as a low probability from any single non-anchor teacher reduces the joint support.
The arithmetic mean retains stronger support from individual teachers and preserves complementary evidence not shared across teachers.
We use the geometric mean to define the teacher-common preferences, and the difference between the two means in log space defines the teacher-specific residual:
\begin{equation}
G(z)=\tfrac{1}{J}\sum\nolimits_k d^k(z),\quad D(z)=\log\bigl(\tfrac{1}{J}\sum\nolimits_k e^{d^k(z)}\bigr)-G(z),
\end{equation}
where $J$ is the number of non-anchor teachers.
The teacher-common preference $G(z)$ is the log ratio of the unnormalized geometric mean to the Answer Teacher distribution.
The teacher-specific residual $D(z)$ is the log ratio of the arithmetic mean to the unnormalized geometric mean, capturing additional support from individual teachers.
Since this support may reflect complementary evidence or conflicting preferences, we gate the residual using directional agreement and relative magnitude:
\begin{equation}
C(z)=\frac{|G(z)|}{\frac{1}{J}\sum_k|d^k(z)|+\epsilon},
\qquad
H(z)=\frac{|G(z)|}{|G(z)|+D(z)+\epsilon},
\end{equation}
where $\epsilon$ is a numerical stability constant.
The agreement gate $C(z)$ reduces the residual contribution when the evidence-specific preferences $d^k(z)$ exhibit opposing signs.
The magnitude gate $H(z)$ limits this contribution relative to $|G(z)|$.
We add the gated residual to the teacher-common preference $G(z)$, subsequently reweighting and normalizing the Answer Teacher distribution:
\begin{equation}
f(z)=G(z)+C(z)H(z)D(z),\qquad
\bar q^{\mathrm{mv}}(z)
=\frac{\bar q^{\mathrm A}(z)e^{f(z)}}
{\sum_u\bar q^{\mathrm A}(u)e^{f(u)}}.
\end{equation}
The fused distribution combines teacher-common preferences with gated teacher-specific residuals from available teachers. When semantic answers are unavailable, an unprivileged anchor teacher is used, and its distribution is retained if no auxiliary teacher is available.

\subsection{Verification-Guided Optimization}

Token-level advantage refinement preserves the signs of
reward-derived advantages but provides limited corrective
guidance for failed trajectories. 
We therefore apply token-level advantage refinement or failure-conditioned distillation according to trajectory verification outcomes.

\noindent\textbf{Trajectory Verification.}
A task-specific verifier assigns an outcome $o_i$ to each trajectory $i$ based on its required output components $\mathcal{C}_i$.
For each component $c \in \mathcal{C}_i$, we first determine its verification status $v(c)$ (more details in Appendix~\ref{app:trajectory_verification}).
The overall outcome is defined as:
\begin{equation}
o_i =
\begin{cases}
\text{Success}, & \text{if } \forall c \in \mathcal{C}_i, v(c) = \text{correct}, \\
\text{Failure}, & \text{if } \exists c \in \mathcal{C}_i, v(c) = \text{incorrect}, \\
\text{Indeterminate}, & \text{otherwise}.
\end{cases}
\end{equation}
The status $v(c)$ is determined by task-specific rules:
For multiple-choice questions, $v(c)$ is correct if the answer matches the ground truth, and incorrect if it is parseable but wrong.
For temporal localization, $v(c)$ is correct when the temporal intersection over union (tIoU) is at least $0.5$, and incorrect when the tIoU is zero.
For temporal multiple-choice questions, both the answer and the interval must be correct.
If a component cannot be parsed, or if the task lacks explicit verification rules, $v(c)$ is marked as unparseable, leading to an \textit{\textbf{Indeterminate}} outcome.
We apply failure-conditioned distillation only to verified failure trajectories that have the required temporal or spatial evidence.

\noindent\textbf{Token-Level Advantage Refinement.}
For non-failure trajectories, we follow RLSD~\cite{yang2026rlsd} and refine token-level advantage magnitudes while preserving their reward-derived signs.
We measure the fused teacher's relative support for each sampled token using the teacher-student log-probability difference:
\begin{equation}
\Delta_{i,t}
=
\operatorname{sg}\!\left[
    \log\bar q_{i,t}^{\mathrm{mv,cr}}(y_{i,t})
    -\log\bar p_{i,t}^{\mathrm{cr}}(y_{i,t})
\right],
\end{equation}
where $y_{i,t}$ is the token at position $t$ in trajectory $i$.
The fused teacher distribution $\bar q_{i,t}^{\mathrm{mv,cr}}$ and student distribution $\bar p_{i,t}^{\mathrm{cr}}$ are conditioned on the same input and student-generated prefix $y_{i,<t}$.
Both distributions are normalized over the shared candidate set, and the operator $\operatorname{sg}$ denotes the stop-gradient operation.
We use $\Delta_{i,t}$ to rescale the trajectory advantage at each token:
\begin{equation}
\begin{aligned}
w_{i,t}
&= \exp\!\bigl(\operatorname{sign}(A_i)\Delta_{i,t}\bigr),\\
A_{i,t}^{\mathrm{cr}}
&= A_i \biggl[
    (1-\lambda_s)
    +\lambda_s\operatorname{clip}
    \Bigl(w_{i,t},\, 1-\epsilon_w,\, 1+\epsilon_w\Bigr)
\biggr].
\end{aligned}
\end{equation}
where $A_i$ is the group-relative advantage of trajectory $i$, $w_{i,t}$ is the token-specific weight, and $A_{i,t}^{\mathrm{cr}}$ is the refined token advantage.
At training step $s$, $\lambda_s\in[0,1]$ controls the strength of teacher reweighting, while $\epsilon_w\in[0,1)$ defines the weight clipping range.
We substitute $A_{i,t}^{\mathrm{cr}}$ into the GRPO clipped objective:
\begin{equation}
\begin{gathered}
\ell_i^{\mathrm{cr}}
=-\tfrac{1}{T_i}\sum\nolimits_{t=1}^{T_i}
\min\!\bigl[
\rho_{i,t}A_{i,t}^{\mathrm{cr}},
\operatorname{clip}(\rho_{i,t},1-\epsilon_{\mathrm{pg}},1+\epsilon_{\mathrm{pg}})
A_{i,t}^{\mathrm{cr}}
\bigr],\\[2pt]
\rho_{i,t}
=\pi_\theta(y_{i,t}\mid x_i,y_{i,<t})
\,/\,
\pi_{\mathrm{old}}(y_{i,t}\mid x_i,y_{i,<t}).
\end{gathered}
\end{equation}
where $\rho_{i,t}$ is the probability ratio between the current and rollout policies.
Here, $T_i$ denotes the number of valid generated tokens, and $\epsilon_{\mathrm{pg}}$ is the policy clipping threshold.

\noindent\textbf{Failure-Conditioned Distillation.}
For the verified failure trajectories, we minimize the reverse KL divergence from the student to the fused teacher distribution, conditioned on the same student-generated prefixes. This reverse-KL loss replaces the GRPO policy loss for these trajectories:
\begin{equation}
\ell_i^{\mathrm{rkl}}
=
\frac{1}{T_i}\sum_{t=1}^{T_i}
D_{\mathrm{KL}}\!\left(
    \bar p_{i,t}^{\mathrm{rkl}}
    \,\middle\|\,
    \operatorname{sg}
    [\widehat q_{i,t}^{\mathrm{rkl}}]
\right),
\end{equation}
where $\bar p_{i,t}^{\mathrm{rkl}}$ is the student distribution and $\widehat q_{i,t}^{\mathrm{rkl}}$ is the fused teacher distribution. Both distributions are defined over the same candidate set, and the probabilities of all other tokens are combined into a single category. During optimization, the teacher target distribution is detached, so gradients flow only through the student distribution. Importantly, this loss does not depend on the advantage $A_i$. It provides corrective supervision even when group-relative advantages vanish, since the reverse KL divergence inherently penalizes any discrepancy between the student and teacher distributions.

\noindent\textbf{Unified Training Objective.}
The overall objective is:
\begin{equation}
    \mathcal L_s(\theta)
    =
    \frac{1}{|\mathcal B_s|}
    \sum_{i\in\mathcal B_s}
    \left[
        (1-g_i)\ell_i^{\mathrm{cr}}
        +g_i\gamma_s\ell_i^{\mathrm{rkl}}
    \right],
\end{equation}
where $\mathcal B_s$ is the valid trajectory batch at step $s$ and $\gamma_s\geq0$ weights the distillation loss. We set $g_i=1$ for trajectories that use failure-conditioned distillation with the self-teacher, and $g_i=0$ otherwise.
Teacher targets remain detached throughout training.
We gradually decay $\lambda_s$ during self-distillation, then set $\lambda_s=g_i=0$ to recover GRPO.
During self-distillation, the shared teacher parameters are initialized as $\bar\theta_0=\theta_0$ and updated after each student update:
\begin{equation}
    \bar\theta_s
    =\beta\bar\theta_{s-1}+(1-\beta)\theta_s,
    \qquad 0\leq\beta<1,
\end{equation}
where $\beta$ is the EMA decay coefficient.
We gradually decay $\lambda_s$ during self-distillation and recover GRPO afterward by setting $\lambda_s=g_i=0$ for all trajectories.

\section{Experiments}
\label{sec:experiments}

\subsection{Experimental Setup}
\label{sec:experimental_setup}

\textbf{Implementation Details.}
MeSD is initialized from Open-o3-Video-SFT-7B and trained on
STGR-RL~\citep{meng2025openo3video} for one epoch.
For a fair comparison, we use the same training data and annotations,
with existing answer and grounding labels serving as teacher
privileged information.
We generate four on-policy responses per prompt using vLLM
at temperature $1.0$, with a maximum response length of 768 tokens.
Videos are sampled at 2 FPS with at most 16 frames.
The learning rate is $10^{-6}$, and token weights are clipped
to $[0.8,1.2]$.
The teacher reweighting coefficient $\lambda_s$ decays linearly
from $0.5$ to zero over the first 1000 steps.
Eligible failures use reverse-KL distillation in place of
the GRPO policy loss, with $\gamma_s$ increasing linearly
from zero to $0.8$ over the first 100 steps.
After step 1000, teacher supervision is disabled and training
continues with standard GRPO.
Further experimental details are provided in
Appendix~\ref{app:evaluation}.

\subsection{Main Results}
\label{sec:main_results}

\begin{table}[t]
\setlength{\belowcaptionskip}{5pt}
\centering
\caption{Performance comparison on V-STAR.
Chain1 and Chain2 denote the what--when--where and what--where--when reasoning orders, respectively.
Best and second-best results are shown in \textbf{bold} and \underline{underlined}.}
\label{tab:vstar}
\small
\renewcommand{\arraystretch}{1.05}
\setlength{\tabcolsep}{3pt}

\begin{adjustbox}{max width=\linewidth}
\begin{NiceTabular}{@{}lccccccc@{}}
\CodeBefore
\rowcolor{orange!10}{3,6,10}
\rowcolor{blue!10}{17}
\rowcolor{gray!15}{18}
\Body
\toprule
\multirow{2}{*}[-2.7pt]{\textbf{Model}}
& \textbf{What}
& \multicolumn{2}{c}{\textbf{When (temporal IoU)}}
& \multicolumn{2}{c}{\textbf{Where (spatial IoU)}}
& \multicolumn{2}{c}{\textbf{Overall}} \\
\cmidrule(lr){3-4}
\cmidrule(lr){5-6}
\cmidrule(lr){7-8}
& \textbf{Acc.}
& \textbf{Chain1} & \textbf{Chain2}
& \textbf{Chain1} & \textbf{Chain2}
& \textbf{mAM} & \textbf{mLGM} \\
\midrule

\multicolumn{8}{l}{\textit{Closed-source Models}} \\
Gemini-2-Flash  ~\citep{google2025gemini2flash}
& 53.0
& 24.5 & 23.8
& 4.6 & 2.2
& 26.9 & 35.6 \\
GPT-4o ~\citep{openai2024gpt4o}
& 60.8
& 16.7 & 12.8
& 6.5 & 3.0
& 26.8 & 38.2 \\

\multicolumn{8}{l}{\textit{Open-source General Models }} \\
Oryx-1.5-7B ~\citep{liu2025oryx}
& 20.5
& 13.5 & 14.8
& 10.1 & 3.5
& 15.1 & 13.8 \\
InternVL-2.5-8B ~\citep{chen2024internvl25}
& 44.2
& 8.7 & 7.8
& 0.7 & 0.1
& 17.6 & 24.9 \\
Qwen2.5-VL-7B ~\citep{bai2025qwen25vl}
& 33.5
& 15.4 & 13.8
& 17.0 & 2.5
& 19.3 & 22.4 \\

\multicolumn{8}{l}{\textit{Video-specialized Models}} \\
TRACE ~\citep{guo2025trace}
& 17.6
& 19.1 & 17.1
& 0.0 & 0.0
& 12.0 & 13.3 \\
VideoLLaMA3-7B ~\citep{zhang2025videollama3}
& 41.9
& 23.0 & 23.1
& 0.9 & 0.2
& 21.7 & 27.0 \\
LLaVA-Video ~\citep{zhang2025llavavideo}
& 49.5
& 10.5 & 12.2
& 1.9 & 1.3
& 20.8 & 27.3 \\
VideoChat2 ~\citep{li2024mvbench}
& 36.2
& 13.7 & 12.5
& 2.5 & 1.0
& 17.0 & 20.3 \\
Open-o3-Video-7B ~\citep{meng2025openo3video}
& 61.0
& 24.5 & 24.0
& 25.4 & 6.0
& 33.7 & 46.6 \\
VISD-7B ~\citep{lin2026visd}
& \underline{61.9}
& \textbf{26.8} & \textbf{27.2}
& \underline{26.7} & \underline{6.3}
& \underline{35.1} & \underline{48.9} \\

\textbf{MeSD-7B (Ours)}
& \textbf{63.7}
& \underline{25.8} & \underline{25.0}
& \textbf{28.1} & \textbf{6.5}
& \textbf{35.5} & \textbf{50.2} \\
$\Delta$ vs. Qwen2.5-VL-7B
& +30.2
& +10.4 & +11.2
& +11.1 & +4.0
& +16.2 & +27.8 \\
\bottomrule
\end{NiceTabular}
\end{adjustbox}

\end{table}


\begin{table*}[htbp]
\setlength{\belowcaptionskip}{5pt}
\centering
\caption{Performance comparison on WorldSense, VideoMMMU, LRR, and TVGBench.
LRR denotes LongVideo-Reason-eval.
Avg denotes the average of WorldSense Overall, VideoMMMU Overall,
LRR Acc., and TVGBench mIoU.
Best and second-best results among open-source models are shown in
\textbf{bold} and \underline{underlined}, respectively.}
\label{tab:general}
\small
\renewcommand{\arraystretch}{1.05}
\setlength{\tabcolsep}{5pt}

\begin{adjustbox}{max width=\linewidth}
\begin{NiceTabular}{@{}lccccccc@{}}
\CodeBefore
\rowcolor{blue!10}{10}
\rowcolor{gray!15}{11}
\Body
\toprule
\multirow{2}{*}[-2.7pt]{\textbf{Model}}
& \multicolumn{2}{c}{\textbf{WorldSense}}
& \multicolumn{2}{c}{\textbf{VideoMMMU}}
& \textbf{LRR}
& \textbf{TVGBench}
& \multirow{2}{*}[-2.7pt]{\textbf{Avg}} \\
\cmidrule(lr){2-3}
\cmidrule(lr){4-5}
\cmidrule(lr){6-6}
\cmidrule(lr){7-7}
& \textbf{Overall}
& \textbf{Understanding}
& \textbf{Overall}
& \textbf{Perception}
& \textbf{Acc.}
& \textbf{mIoU}
& \\
\midrule

GPT-4o
& 42.6 & --
& 61.2 & 66.0
& -- & -- & -- \\
\midrule

Qwen2.5-VL-7B ~\citep{bai2025qwen25vl}
& 35.5 & 38.4
& 51.2 & 64.7
& 59.3 & 16.3 & 40.6 \\

VideoRFT-7B ~\citep{wang2025videorft}
& 38.4 & 40.0
& 51.1 & 66.0
& 69.4 & 14.3 & 43.3 \\

VideoR1-7B ~\citep{feng2025videor1}
& 36.9 & 37.7
& 52.4 & 65.3
& 68.9 & 9.6 & 42.0 \\

Open-o3-Video-7B ~\citep{meng2025openo3video}
& 38.9 & 39.7
& 52.6 & 67.4
& 69.1 & 19.2 & 45.0 \\

VisionCoach-7B ~\citep{lee2026visioncoach}
& 39.2 & 40.2
& \underline{53.8} & \underline{70.3}
& 70.7 & 21.0 & 46.2 \\

VISD-7B ~\citep{lin2026visd}
& \underline{39.9} & \underline{40.4}
& \underline{53.8} & 69.3
& \underline{73.5} & \underline{24.8}
& \underline{48.0} \\
\midrule

\textbf{MeSD-7B (Ours)}
& \textbf{41.6} & \textbf{46.0}
& \textbf{56.1} & \textbf{73.0}
& \textbf{73.6} & \textbf{29.8}
& \textbf{50.3} \\

$\Delta$ vs. Qwen2.5-VL-7B
& +6.1 & +7.6
& +4.9 & +8.3
& +14.3 & +13.5 & +9.7 \\
\bottomrule
\end{NiceTabular}
\end{adjustbox}

\end{table*}

\textbf{What, When, and Where: Grounded Reasoning on V-STAR.}
Table~\ref{tab:vstar} demonstrates that MeSD substantially outperforms Qwen2.5-VL-7B across all evaluation dimensions. Answer accuracy achieves the largest absolute gain, increasing from 33.5\% to 63.7\%. Meanwhile, improvements in temporal and spatial IoU indicate that these more accurate answers are accompanied by better event and object localization. Compared to VISD, the advantages of MeSD lie primarily in answer accuracy and spatial grounding, achieving the best results among all evaluated models in both dimensions. Although its temporal IoU is slightly lower than that of VISD, MeSD achieves the highest aggregate scores, with an mAM of 35.5\% and an mLGM of 50.2\%. This highlights its ability to combine accurate answers with explicit visual grounding.

\textbf{Performance across Diverse Video Tasks.}
As shown in Table~\ref{tab:general}, MeSD achieves the highest performance among the evaluated open-source models across all reported metrics. Compared to Qwen2.5-VL-7B, MeSD increases the average score from 40.6\% to 50.3\%. The most substantial benchmark-level gains occur on LRR and TVGBench, indicating stronger long-video reasoning and more precise temporal localization. Furthermore, MeSD outperforms VISD on all four benchmarks, increasing the average score from 48.0\% to 50.3\%. The most notable improvements are observed in temporal grounding, WorldSense understanding, and VideoMMMU perception. These advancements over a strong self-distillation baseline validate the effectiveness of MeSD.

\Needspace{190pt}
\begingroup
\setlength{\intextsep}{4pt}
\setlength{\columnsep}{12pt}
\begin{wraptable}{r}{0.5\textwidth}
\centering
\captionsetup{skip=2pt}

\caption{\textit{Charades-STA} results. Best and second-best results are
shown in \textbf{bold} and \underline{underlined}.}
\label{tab:charades_sta}

\scriptsize
\setlength{\tabcolsep}{2pt}
\renewcommand{\arraystretch}{1.05}

\begin{adjustbox}{max width=\linewidth}
\begin{tabular*}{\linewidth}{
    @{\extracolsep{\fill}}lcccc@{}
}
\toprule
\textbf{Model}
& \textbf{R@0.3}
& \textbf{R@0.5}
& \textbf{R@0.7}
& \textbf{mIoU} \\
\midrule

\rowcolor{orange!10}
\multicolumn{5}{l}{\textit{General Video LLMs}} \\
VideoChat ~\citep{li2023videochat}       & 9.0  & 3.3  & 1.3  & 6.5  \\
VideoLLaMA ~\citep{zhang2023videollama}      & 10.4 & 3.8  & 0.9  & 7.1  \\
Video-ChatGPT ~\citep{maaz2024videochatgpt}   & 20.0 & 7.7  & 1.7  & 13.7 \\
Valley ~\citep{luo2023valley}          & 28.4 & 1.8  & 0.3  & 21.4 \\

\rowcolor{orange!10}
\multicolumn{5}{l}{\textit{Temporal Grounding Video LLMs}} \\
TimeChat ~\citep{ren2024timechat}        & --   & 32.2 & 13.4 & --   \\
HawkEye ~\citep{wang2024hawkeye}         & 50.6 & 31.4 & 14.5 & 33.7 \\
VTG-LLM ~\citep{guo2025vtgllm}         & 51.2 & 33.8 & 15.7 & 34.4 \\
VTimeLLM ~\citep{huang2024vtimellm}        & 51.0 & 27.5 & 11.4 & 31.2 \\

\rowcolor{orange!10}
\multicolumn{5}{l}{\textit{Grounded Reasoning Models}} \\
Open-o3-video ~\citep{meng2025openo3video}
& 62.6
& 45.6
& \underline{24.5}
& 42.5 \\

VISD ~\citep{lin2026visd}
& \underline{67.3}
& \underline{46.1}
& 21.6
& \underline{44.6} \\

\textbf{MeSD (Ours)}
& \textbf{73.1}
& \textbf{54.7}
& \textbf{28.8}
& \textbf{49.2} \\

\bottomrule
\end{tabular*}
\end{adjustbox}
\end{wraptable}

\textbf{Zero-Shot Temporal Grounding on Charades-STA.}
Table~\ref{tab:charades_sta} presents the zero-shot temporal grounding results in terms of recall at IoU thresholds of 0.3, 0.5, and 0.7, alongside mIoU. 
MeSD outperforms all compared methods across these four metrics. 
Compared to VISD, it improves R@0.5 and R@0.7 by 8.6\% and 7.2\%, respectively, and increases mIoU from 44.6\% to 49.2\%. 
Furthermore, at the highest IoU threshold, MeSD surpasses Open-o3-Video, an achievement VISD fails to match. 
These results demonstrate more precise temporal alignment and validate the transferability of the evidence-based supervision in MeSD.
\par
\ifnum\value{WF@wrappedlines}>1
  \vspace{\dimexpr\value{WF@wrappedlines}\baselineskip-\baselineskip\relax}
\fi
\WFclear
\endgroup

\subsection{Ablation Studies and Analysis}
\label{sec:ablation_studies}

\begin{table}[!htbp]
\setlength{\belowcaptionskip}{5pt}
\centering
\caption{Ablation study on MeSD components and training strategies.
FCD denotes failure-conditioned distillation, ES denotes evidence separation, T\&S denotes temporal and spatial evidence, GPA denotes gated preference aggregation, and VGO denotes verification-guided optimization. 
The best results in each column are highlighted in bold. More details in Appendix~\ref{app:detailed_results}.}
\label{tab:ablation}
\small
\renewcommand{\arraystretch}{1.05}
\setlength{\tabcolsep}{3pt}

\begin{adjustbox}{max width=\linewidth}
\begin{NiceTabular}{@{}lcccccccc@{}}
\CodeBefore
\rowcolor{orange!12}{3,8}
\rowcolor{blue!10}{13}
\Body
\toprule
\multirow{2}{*}[-2.7pt]{\textbf{Model}}
& \multicolumn{2}{c}{\textbf{V-STAR}}
& \multicolumn{2}{c}{\textbf{WorldSense}}
& \multicolumn{2}{c}{\textbf{Video-MMMU}}
& \textbf{LRR}
& \textbf{TVGBench} \\
\cmidrule(lr){2-3}
\cmidrule(lr){4-5}
\cmidrule(lr){6-7}
\cmidrule(lr){8-8}
\cmidrule(lr){9-9}
& mAM
& mLGM
& Overall
& Understanding
& Overall
& Perception
& Acc.
& mIoU \\
\midrule

\multicolumn{9}{l}{\textit{Component Variants}} \\
w/o FCD
& 30.7
& 41.0
& 38.6
& 38.7
& 53.8
& 68.3
& 71.3
& 23.0 \\
w/o ES
& 30.9
& 41.3
& 38.3
& 38.4
& 55.4
& 71.0
& 70.2
& 22.4 \\
w/o T\&S
& 30.8
& 41.2
& 37.9
& 39.3
& 54.8
& 71.0
& 70.6
& 22.3 \\
w/o GPA
& 30.9
& 41.4
& 39.5
& 39.5
& 53.9
& 69.7
& 70.4
& 22.9 \\

\multicolumn{9}{l}{\textit{Training Strategy Variants}} \\
Base VLM
& 19.3
& 22.4
& 35.5
& 38.4
& 51.2
& 64.7
& 59.3
& 16.3 \\
GSPO
& 33.7
& 46.6
& 37.5
& 39.7
& 52.3
& 68.0
& 69.4
& 20.8 \\
GRPO
& 30.6
& 40.7
& 39.2
& 40.4
& 53.2
& 69.3
& 71.9
& 22.5 \\
w/o VGO
& 30.6
& 40.9
& 39.5
& 41.0
& 54.1
& 67.7
& 71.2
& 20.7 \\
\midrule

\textbf{MeSD}
& \textbf{35.5}
& \textbf{50.2}
& \textbf{41.6}
& \textbf{46.0}
& \textbf{56.1}
& \textbf{73.0}
& \textbf{73.6}
& \textbf{29.8} \\
\bottomrule
\end{NiceTabular}
\end{adjustbox}

\end{table}

\textbf{Effects of Evidence Separation and Complementarity.}
Table~\ref{tab:ablation} compares MeSD with two ablated variants: a single teacher conditioned on all available evidence (w/o ES) and an answer-only teacher (w/o T\&S). 
Both variants underperform compared to MeSD across all metrics. Notably, the TVGBench mIoU drops from 29.8\% to 22.4\% and 22.3\%, respectively. 
These results support incorporating temporal and spatial evidence and modeling their preferences separately.

\textbf{Effects of Gated Preference Aggregation.}
With the evidence-conditioned teachers fixed, we replace the gated preference aggregation with a simple arithmetic averaging of the evidence-specific preferences (w/o GPA). 
This modification degrades performance across all benchmarks (Table~\ref{tab:ablation}). 
Specifically, V-STAR mLGM drops from 50.2\% to 41.4\%, and TVGBench mIoU decreases from 29.8\% to 22.9\%. 
These results demonstrate the advantages of selectively retaining teacher-specific support while suppressing conflicting contributions.

\textbf{Effects of Failure-Conditioned Distillation.}
Applying token-level advantage refinement to all trajectories without FCD reduces the V-STAR mLGM and TVGBench mIoU by 9.2\% and 6.8\%, respectively (Table~\ref{tab:ablation}). 
The training curves exhibit a similar pattern: the model without FCD reaches a lower reward plateau, whereas the MeSD reward continues to increase before stabilizing (Figure~\ref{train_curve}(a)). 
These results indicate that the corrective guidance provided by advantage refinement for failed trajectories is inherently limited. 
Failure-conditioned distillation employs the fused teacher distribution for reverse-KL correction, thereby providing supervision over candidate tokens that extends beyond sampled-token reweighting.

\textbf{Effects of Optimization Strategy.}
MeSD outperforms GRPO and GSPO across all reported metrics (Table~\ref{tab:ablation}), raising V-STAR mLGM from GSPO's 46.6\% to 50.2\% and TVGBench mIoU from GRPO's 22.5\% to 29.8\%.
These gains support complementing reward-based optimization with evidence-conditioned teacher supervision for spatio-temporal reasoning and general video tasks.

\textbf{Effects of Verification-Guided Optimization.}
Replacing VGO with vanilla OPSD (w/o VGO) while retaining the multi-evidence teachers lowers all reported metrics, with TVGBench mIoU decreasing from 29.8\% to 20.7\% (Table~\ref{tab:ablation}).
Figure~\ref{train_curve}(b) further shows declining training rewards without VGO, whereas MeSD improves and stabilizes at a higher level.
These results suggest that multi-evidence supervision alone is insufficient, supporting verification-guided teacher supervision.

\begin{figure}[tp]
    \centering
    \includegraphics[width=0.92\textwidth]{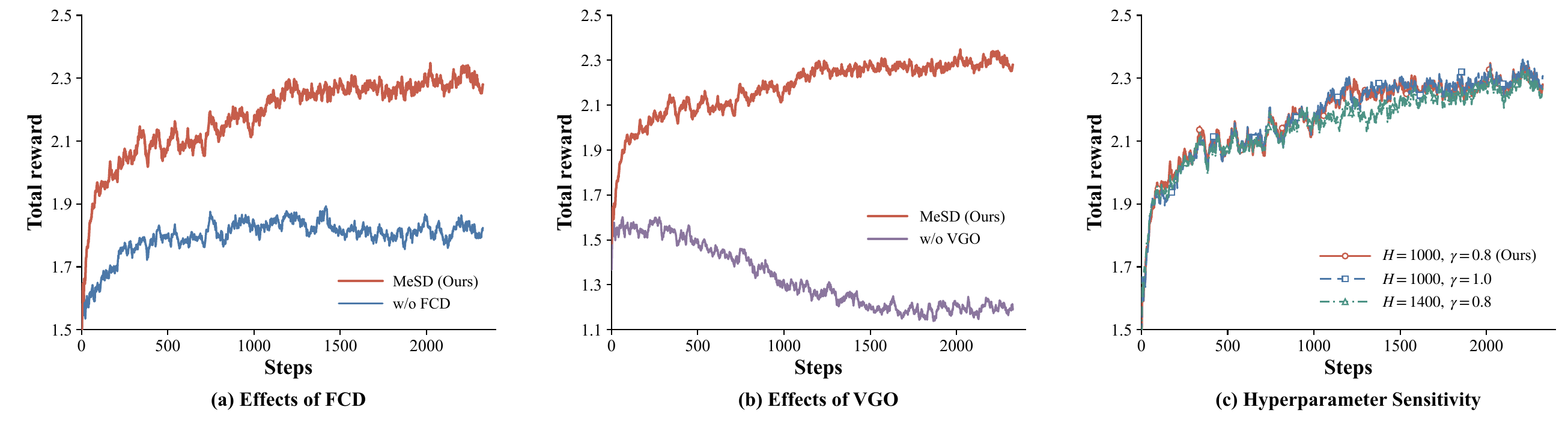}
    \caption{Training ablation curves. MeSD achieves higher training rewards than (a) credit refinement alone (w/o FCD) and (b) multi-evidence OPSD (w/o VGO). (c) Varying the teacher cutoff step $H$ and $\gamma$ yields similar reward trajectories.}
    \label{train_curve}
\end{figure}

\begin{figure}[tp]
    \centering
    \includegraphics[width=0.92\textwidth]{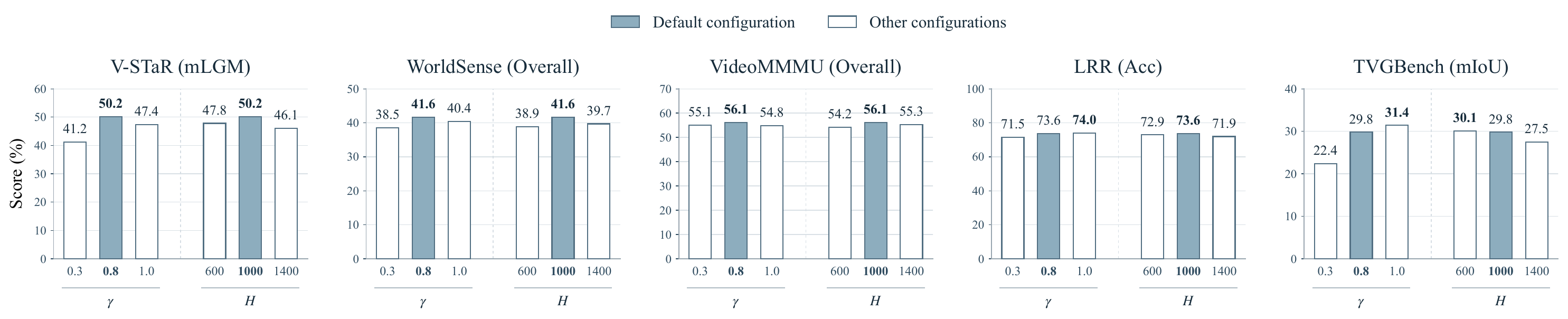}
    \caption{Sensitivity of MeSD to $\gamma$ and the teacher cutoff step $H$ across five benchmarks. Filled bars indicate the default settings ($\gamma=0.8$, $H=1000$).}
    \label{gamma}
\end{figure}

\textbf{Effects of Correction Strength and Teacher Guidance Duration.}
We separately vary $\gamma\in\{0.3,0.8,1.0\}$ (direct-correction strength after warm-up) and $H\in\{600,1000,1400\}$ (teacher-supervision cutoff), keeping the other parameter at its default.
The three configurations in Figure~\ref{train_curve}(c) exhibit similar reward trajectories and final rewards, suggesting training robustness within the tested ranges.
Figure~\ref{gamma} shows that $\gamma=0.8$ and $H=1000$ perform best on three and four benchmarks, respectively, supporting these defaults as a balanced choice across tasks.

\section{Conclusion}
\label{sec:conclusion}

We introduced MeSD, a multi-evidence self-distillation framework for VideoLLMs that combines evidence-aware supervision with verification-guided optimization. Using an answer-conditioned teacher as a semantic anchor, MeSD integrates temporal and spatial evidence through teacher-common preferences and gated teacher-specific residuals. It preserves reward-derived advantage signs during credit refinement and reserves reverse-KL correction for verified failures with the required evidence. Experiments demonstrate overall gains over evaluated baselines across video understanding, spatiotemporal reasoning, and temporal grounding, including improvements across Qwen3-VL model sizes. Ablations support evidence separation, gated fusion, and selective correction for effective video post-training using existing annotations.
\clearpage

\clearpage
\endgroup
\bibliographystyle{plainnat}
\bibliography{paper}

@inproceedings{feng2025videor1,
  title     = {{Video-R1}: Reinforcing Video Reasoning in {MLLMs}},
  author    = {Feng, Kaituo and Gong, Kaixiong and Li, Bohao and
               Guo, Zonghao and Wang, Yibing and Peng, Tianshuo and
               Wu, Junfei and Zhang, Xiaoying and Wang, Benyou and
               Yue, Xiangyu},
  booktitle = {Advances in Neural Information Processing Systems},
  volume    = {38},
  year      = {2025}
}

@inproceedings{wang2025timer1,
  title     = {{Time-R1}: Post-Training Large Vision Language Model
               for Temporal Video Grounding},
  author    = {Wang, Ye and Wang, Ziheng and Xu, Boshen and Du, Yang and
               Lin, Kejun and Xiao, Zihan and Yue, Zihao and
               Ju, Jianzhong and Zhang, Liang and Yang, Dingyi and
               Fang, Xiangnan and He, Zewen and Luo, Zhenbo and
               Wang, Wenxuan and Lin, Junqi and Luan, Jian and Jin, Qin},
  booktitle = {Advances in Neural Information Processing Systems},
  volume    = {38},
  year      = {2025}
}

@article{li2026videoopd,
  title   = {{Video-OPD}: Efficient Post-Training of Multimodal
             Large Language Models for Temporal Video Grounding
             via On-Policy Distillation},
  author  = {Li, Jiaze and Yin, Hao and Xu, Haoran and Xu, Boshen and
             Tan, Wenhui and He, Zewen and Ju, Jianzhong and
             Luo, Zhenbo and Luan, Jian},
  journal = {arXiv preprint arXiv:2602.02994},
  year    = {2026}
}

@article{meng2025openo3video,
  title   = {{Open-o3-Video}: Grounded Video Reasoning with
             Explicit Spatio-Temporal Evidence},
  author  = {Meng, Jiahao and Li, Xiangtai and Wang, Haochen and
             Tan, Yue and Zhang, Tao and Kong, Lingdong and
             Tong, Yunhai and Wang, Anran and Teng, Zhiyang and
             Wang, Yujing and Wang, Zhuochen},
  journal = {arXiv preprint arXiv:2510.20579},
  year    = {2025}
}

@inproceedings{agarwal2024gkd,
  title     = {On-Policy Distillation of Language Models:
               Learning from Self-Generated Mistakes},
  author    = {Agarwal, Rishabh and Vieillard, Nino and Zhou, Yongchao and
               Stanczyk, Piotr and Ramos Garea, Sabela and
               Geist, Matthieu and Bachem, Olivier},
  booktitle = {International Conference on Learning Representations},
  volume    = {2024},
  pages     = {21246--21263},
  year      = {2024},
}

@article{zhao2026opsd,
  title   = {Self-Distilled Reasoner: On-Policy Self-Distillation
             for Large Language Models},
  author  = {Zhao, Siyan and Xie, Zhihui and Liu, Mengchen and
             Huang, Jing and Pang, Guan and Chen, Feiyu and Grover, Aditya},
  journal = {arXiv preprint arXiv:2601.18734},
  year    = {2026}
}

@article{yang2026rlsd,
  title   = {Self-Distilled {RLVR}},
  author  = {Yang, Chenxu and Qin, Chuanyu and Si, Qingyi and
             Chen, Minghui and Gu, Naibin and Yao, Dingyu and
             Lin, Zheng and Wang, Weiping and Wang, Jiaqi and Duan, Nan},
  journal = {arXiv preprint arXiv:2604.03128},
  year    = {2026}
}

@article{nguyen2026avsd,
  title   = {{AVSD}: Adaptive-View Self-Distillation by Balancing
             Consensus and Teacher-Specific Privileged Signals},
  author  = {Nguyen, Duy and Xiao, Hanqi and Prasad, Archiki and
             Khan, Zaid and Das, Anirban and Zhang, Austin and
             Sahu, Sambit and Lee, Hyunji and Stengel-Eskin, Elias and
             Bansal, Mohit},
  journal = {arXiv preprint arXiv:2605.20643},
  year    = {2026}
}

@article{cai2026h2sd,
  title   = {{H$^2$SD}: Hybrid Hindsight Self-Distillation},
  author  = {Cai, Qiye and Ma, Yichuan and Li, Peiji and
             Chen, Yongkang and Guo, Qipeng and Zou, Yicheng and
             Li, Linyang and Feng, Xiaocheng and Qin, Bing},
  journal = {arXiv preprint arXiv:2607.18955},
  year    = {2026}
}

@article{yuan2026visionopd,
  title   = {{Vision-OPD}: Learning to See Fine Details for
             Multimodal {LLMs} via On-Policy Self-Distillation},
  author  = {Yuan, Qianhao and Lou, Jie and Yu, Xing and
             Lin, Hongyu and Sun, Le and Han, Xianpei and Lu, Yaojie},
  journal = {arXiv preprint arXiv:2605.18740},
  year    = {2026}
}

@article{wang2026videoopsd,
  title   = {{Video-OPSD}: Exploiting Privileged Visual Evidence
             for On-Policy Self-Distillation in Video Large Language Models},
  author  = {Wang, Ziyue and Huang, Shiqi and Xu, Weiwen and
             Wen, Bihan and Jiang, Xudong},
  journal = {arXiv preprint arXiv:2608.27065},
  year    = {2026}
}

@article{venkatraman2026s3t,
  title   = {Temporal Self-Distillation: Learning Visual State
             Tracking in Videos Without Supervision},
  author  = {Venkatraman, Shravan and Zhao, Wenshuai and
             Vali, Mohammad Hassan and Solin, Arno},
  journal = {arXiv preprint arXiv:2609.04203},
  year    = {2026}
}

@article{lin2026visd,
  title   = {{VISD}: Enhancing Video Reasoning via Structured Self-Distillation},
  author  = {Lin, Hao and Lv, Kunyang and Jiang, Xu and
             Tian, Jingqi and Du, Zhongjing and Ding, Jiayu and
             Zhang, Qiaoman and Jin, Hongbo},
  journal = {arXiv preprint arXiv:2605.06094},
  year    = {2026}
}

@inproceedings{wang2025videorft,
  title     = {{VideoRFT}: Incentivizing Video Reasoning Capability
               in {MLLMs} via Reinforced Fine-Tuning},
  author    = {Wang, Qi and Yu, Yanrui and Yuan, Ye and
               Mao, Rui and Zhou, Tianfei},
  booktitle = {Advances in Neural Information Processing Systems},
  volume    = {38},
  pages     = {4350--4376},
  publisher = {Curran Associates, Inc.},
  year      = {2025},
}

@inproceedings{park2025deepvideor1,
  title     = {{DeepVideo-R1}: Video Reinforcement Fine-Tuning
               via Difficulty-aware Regressive {GRPO}},
  author    = {Park, Jinyoung and Na, Jeehye and Kim, Jinyoung and
               Kim, Hyunwoo J.},
  booktitle = {Advances in Neural Information Processing Systems},
  volume    = {38},
  pages     = {153710--153737},
  year      = {2025},
  doi       = {10.52202/085713-4626}
}

@article{li2025videochatr1,
  title   = {{VideoChat-R1}: Enhancing Spatio-Temporal Perception
             via Reinforcement Fine-Tuning},
  author  = {Li, Xinhao and Yan, Ziang and Meng, Desen and
             Dong, Lu and Zeng, Xiangyu and He, Yinan and
             Wang, Yali and Qiao, Yu and Wang, Yi and Wang, Limin},
  journal = {arXiv preprint arXiv:2504.06958},
  year    = {2025}
}

@article{lee2026visioncoach,
  title   = {{VisionCoach}: Reinforcing Grounded Video Reasoning
             via Visual-Perception Prompting},
  author  = {Lee, Daeun and Yu, Shoubin and Zhang, Yue and Bansal, Mohit},
  journal = {arXiv preprint arXiv:2603.14659},
  year    = {2026}
}

@InProceedings{Wu_2026_CVPR,
    author    = {Wu, Zhiheng and Wang, Tong and Wang, Shuning and Liu, Naiming and Zhang, Yumeng},
    title     = {See Further, Think Deeper: Advancing VLM's Reasoning Ability with Low-level Visual Cues and Reflection},
    booktitle = {Proceedings of the IEEE/CVF Conference on Computer Vision and Pattern Recognition (CVPR)},
    month     = {June},
    year      = {2026},
    pages     = {18870-18880}
}

@article{guo2026segment,
  title={Segment policy optimization: Effective segment-level credit assignment in rl for large language models},
  author={Guo, Yiran and Xu, Lijie and Liu, Jie and Ye, Dan and Qiu, Shuang},
  journal={Advances in Neural Information Processing Systems},
  volume={38},
  pages={114399--114431},
  year={2025}
}

@article{guo2025deepseek,
  title={Deepseek-r1: Incentivizing reasoning capability in llms via reinforcement learning},
  author={Guo, Daya and Yang, Dejian and Zhang, Haowei and Song, Junxiao and Wang, Peiyi and Zhu, Qihao and Xu, Runxin and Zhang, Ruoyu and Ma, Shirong and Bi, Xiao and others},
  journal={arXiv preprint arXiv:2501.12948},
  year={2025}
}

@article{hubotter2026sdpo,
  title   = {Reinforcement Learning via Self-Distillation},
  author  = {H{\"u}botter, Jonas and L{\"u}beck, Frederike and
             Behric, Lejs and Baumann, Anton and Bagatella, Marco and
             Marta, Daniel and Hakimi, Ido and Shenfeld, Idan and
             Kleine Buening, Thomas and Guestrin, Carlos and
             Krause, Andreas},
  journal = {arXiv preprint arXiv:2601.20802},
  year    = {2026}
}

@article{ma2026mopd,
  title   = {{MOPD}: Multi-Teacher On-Policy Distillation
             for Capability Integration in {LLM} Post-Training},
  author  = {Ma, Wenhan and Wei, Jianyu and Zhao, Liang and
             Zhang, Hailin and Xiao, Bangjun and Li, Lei and
             Yang, Qibin and Gao, Bofei and Wang, Yudong and
             Li, Rang and Dong, Jinhao and Sui, Zhifang and Luo, Fuli},
  journal = {arXiv preprint arXiv:2606.30406},
  year    = {2026}
}

@article{he2026mtsdpo,
  title   = {Learn from Whoever Is Right: Answer-Verified
             Multi-Teacher Distillation for Multi-Domain {LLMs}},
  author  = {He, Xixiang and Li, Xingming and Wu, Baiqi and
             Sun, Qiyao and Ji, Xuanyu and Cheng, Ao and Hu, Qingyong},
  journal = {arXiv preprint arXiv:2609.02548},
  year    = {2026}
}

@article{xu2026vgopd,
  title   = {Who Teaches Which Token? Verifier-Gated Multi-Expert
             On-Policy Distillation for Scientific Reasoning},
  author  = {Xu, Xun and Zhang, Zaixi},
  journal = {arXiv preprint arXiv:2609.15404},
  year    = {2026}
}

@misc{google2025gemini2flash,
  author = {{Google DeepMind}},
  title = {{Gemini 2.0 Flash} Model Card},
  year = {2025}
}

@misc{openai2024gpt4o,
  author = {{OpenAI}},
  title = {Hello {GPT-4o}},
  year = {2024}
}

@inproceedings{liu2025oryx,
  title = {{Oryx MLLM}: On-Demand Spatial-Temporal Understanding at Arbitrary Resolution},
  author = {Zuyan Liu and Yuhao Dong and Ziwei Liu and Winston Hu and Jiwen Lu and Yongming Rao},
  booktitle = {International Conference on Learning Representations},
  pages = {85485--85507},
  volume = {2025},
  year = {2025}
}

@article{chen2024internvl25,
  title = {Expanding Performance Boundaries of Open-Source Multimodal Models with Model, Data, and Test-Time Scaling},
  author = {Zhe Chen and Weiyun Wang and Yue Cao and Yangzhou Liu and Zhangwei Gao and Erfei Cui and Jinguo Zhu and Shenglong Ye and Hao Tian and Zhaoyang Liu and Lixin Gu and Xuehui Wang and Qingyun Li and Yiming Ren and Zixuan Chen and Jiapeng Luo and Jiahao Wang and Tan Jiang and Bo Wang and Conghui He and Botian Shi and Xingcheng Zhang and Han Lv and Yi Wang and Wenqi Shao and Pei Chu and Zhongying Tu and Tong He and Zhiyong Wu and Huipeng Deng and Jiaye Ge and Kai Chen and Kaipeng Zhang and Limin Wang and Min Dou and Lewei Lu and Xizhou Zhu and Tong Lu and Dahua Lin and Yu Qiao and Jifeng Dai and Wenhai Wang},
  journal = {arXiv preprint arXiv:2412.05271},
  year = {2024}
}

@article{bai2025qwen25vl,
  title = {{Qwen2.5-VL} Technical Report},
  author = {Shuai Bai and Keqin Chen and Xuejing Liu and Jialin Wang and Wenbin Ge and Sibo Song and Kai Dang and Peng Wang and Shijie Wang and Jun Tang and Humen Zhong and Yuanzhi Zhu and Mingkun Yang and Zhaohai Li and Jianqiang Wan and Pengfei Wang and Wei Ding and Zheren Fu and Yiheng Xu and Jiabo Ye and Xi Zhang and Tianbao Xie and Zesen Cheng and Hang Zhang and Zhibo Yang and Haiyang Xu and Junyang Lin},
  journal = {arXiv preprint arXiv:2502.13923},
  year = {2025}
}

@inproceedings{guo2025trace,
  title = {{TRACE}: Temporal Grounding Video {LLM} via Causal Event Modeling},
  author = {Yongxin Guo and Jingyu Liu and Mingda Li and Qingbin Liu and Xi Chen and Xiaoying Tang},
  booktitle = {International Conference on Learning Representations},
  pages = {89594--89618},
  volume = {2025},
  year = {2025}
}

@article{zhang2025videollama3,
  title = {{VideoLLaMA 3}: Frontier Multimodal Foundation Models for Image and Video Understanding},
  author = {Boqiang Zhang and Kehan Li and Zesen Cheng and Zhiqiang Hu and Yuqian Yuan and Guanzheng Chen and Sicong Leng and Yuming Jiang and Hang Zhang and Xin Li and Peng Jin and Wenqi Zhang and Fan Wang and Lidong Bing and Deli Zhao},
  journal = {arXiv preprint arXiv:2501.13106},
  year = {2025}
}

@article{zhang2025llavavideo,
  title = {{LLaVA-Video}: Video Instruction Tuning with Synthetic Data},
  author = {Yuanhan Zhang and Jinming Wu and Wei Li and Bo Li and Zejun Ma and Ziwei Liu and Chunyuan Li},
  journal = {Transactions on Machine Learning Research},
  year = {2025}
}

@inproceedings{li2024mvbench,
  title = {{MVBench}: A Comprehensive Multi-modal Video Understanding Benchmark},
  author = {Kunchang Li and Yali Wang and Yinan He and Yizhuo Li and Yi Wang and Yi Liu and Zun Wang and Jilan Xu and Guo Chen and Ping Luo and Limin Wang and Yu Qiao},
  booktitle = {Proceedings of the IEEE/CVF Conference on Computer Vision and Pattern Recognition},
  pages = {22195--22206},
  year = {2024}
}

@article{li2023videochat,
  title = {{VideoChat}: Chat-Centric Video Understanding},
  author = {KunChang Li and Yinan He and Yi Wang and Yizhuo Li and Wenhai Wang and Ping Luo and Yali Wang and Limin Wang and Yu Qiao},
  journal = {arXiv preprint arXiv:2305.06355},
  year = {2023}
}

@inproceedings{zhang2023videollama,
  title = {{Video-LLaMA}: An Instruction-tuned Audio-Visual Language Model for Video Understanding},
  author = {Hang Zhang and Xin Li and Lidong Bing},
  booktitle = {Proceedings of the 2023 Conference on Empirical Methods in Natural Language Processing: System Demonstrations},
  pages = {543--553},
  doi = {10.18653/v1/2023.emnlp-demo.49},
  year = {2023}
}

@inproceedings{maaz2024videochatgpt,
  title = {{Video-ChatGPT}: Towards Detailed Video Understanding via Large Vision and Language Models},
  author = {Muhammad Maaz and Hanoona Rasheed and Salman Khan and Fahad Khan},
  booktitle = {Proceedings of the 62nd Annual Meeting of the Association for Computational Linguistics (Volume 1: Long Papers)},
  pages = {12585--12602},
  doi = {10.18653/v1/2024.acl-long.679},
  year = {2024}
}

@article{luo2023valley,
  title = {{Valley}: Video Assistant with Large Language model Enhanced abilitY},
  author = {Ruipu Luo and Ziwang Zhao and Min Yang and Zheming Yang and Minghui Qiu and Tao Wang and Zhongyu Wei and Yanhao Wang and Cen Chen},
  journal = {arXiv preprint arXiv:2306.07207},
  year = {2023}
}

@inproceedings{ren2024timechat,
  title = {{TimeChat}: A Time-sensitive Multimodal Large Language Model for Long Video Understanding},
  author = {Shuhuai Ren and Linli Yao and Shicheng Li and Xu Sun and Lu Hou},
  booktitle = {Proceedings of the IEEE/CVF Conference on Computer Vision and Pattern Recognition},
  pages = {14313--14323},
  year = {2024}
}

@article{wang2024hawkeye,
  title = {{HawkEye}: Training Video-Text {LLMs} for Grounding Text in Videos},
  author = {Yueqian Wang and Xiaojun Meng and Jianxin Liang and Yuxuan Wang and Qun Liu and Dongyan Zhao},
  journal = {arXiv preprint arXiv:2403.10228},
  year = {2024}
}

@inproceedings{guo2025vtgllm,
  title = {{VTG-LLM}: Integrating Timestamp Knowledge into Video {LLMs} for Enhanced Video Temporal Grounding},
  author = {Yongxin Guo and Jingyu Liu and Mingda Li and Dingxin Cheng and Xiaoying Tang and Dianbo Sui and Qingbin Liu and Xi Chen and Kevin Zhao},
  booktitle = {Proceedings of the AAAI Conference on Artificial Intelligence},
  volume = {39},
  number = {3},
  pages = {3302--3310},
  doi = {10.1609/aaai.v39i3.32341},
  year = {2025}
}

@inproceedings{huang2024vtimellm,
  title = {{VTimeLLM}: Empower {LLM} to Grasp Video Moments},
  author = {Bin Huang and Xin Wang and Hong Chen and Zihan Song and Wenwu Zhu},
  booktitle = {Proceedings of the IEEE/CVF Conference on Computer Vision and Pattern Recognition},
  pages = {14271--14280},
  year = {2024}
}

@inproceedings{cheng2026vstar,
  title = {{V-STaR}: Benchmarking Video-{LLMs} on Video Spatio-Temporal Reasoning},
  author = {Zixu Cheng and Jian Hu and Ziquan Liu and Chenyang Si and Wei Li and Shaogang Gong},
  booktitle = {Proceedings of the IEEE/CVF Conference on Computer Vision and Pattern Recognition Findings},
  pages = {9155--9164},
  year = {2026}
}

@inproceedings{hong2026worldsense,
  title = {{WorldSense}: Evaluating Real-world Omnimodal Understanding for Multimodal {LLMs}},
  author = {Jack Hong and Shilin Yan and Jiayin Cai and Xiaolong Jiang and Yao Hu and Weidi Xie},
  booktitle = {International Conference on Learning Representations},
  pages = {52423--52443},
  volume = {2026},
  year = {2026}
}

@article{hu2025videommmu,
  title = {{Video-MMMU}: Evaluating Knowledge Acquisition from Multi-Discipline Professional Videos},
  author = {Kairui Hu and Penghao Wu and Fanyi Pu and Wang Xiao and Yuanhan Zhang and Xiang Yue and Bo Li and Ziwei Liu},
  journal = {arXiv preprint arXiv:2501.13826},
  year = {2025}
}

@inproceedings{chen2025scalingrl,
  title = {Scaling {RL} to Long Videos},
  author = {Yukang Chen and Wei Huang and Baifeng Shi and Qinghao Hu and Hanrong Ye and Ligeng Zhu and Zhijian Liu and Pavlo Molchanov and Jan Kautz and Xiaojuan Qi and Sifei Liu and Hongxu Yin and Yao Lu and Song Han},
  booktitle = {Advances in Neural Information Processing Systems},
  volume = {38, Main Conference},
  pages = {172842--172870},
  doi = {10.52202/085713-5753},
  year = {2025}
}

@inproceedings{gao2017tall,
  title = {{TALL}: Temporal Activity Localization via Language Query},
  author = {Jiyang Gao and Chen Sun and Zhenheng Yang and Ram Nevatia},
  booktitle = {Proceedings of the IEEE International Conference on Computer Vision},
  pages = {5267--5275},
  year = {2017}
}

@article{bai2025qwen3vl,
  title = {{Qwen3-VL} Technical Report},
  author = {Shuai Bai and Yuxuan Cai and Ruizhe Chen and Keqin Chen
    and Xionghui Chen and Zesen Cheng and Lianghao Deng and Wei Ding
    and Chang Gao and Chunjiang Ge and Wenbin Ge and Zhifang Guo
    and Qidong Huang and Jie Huang and Fei Huang and Binyuan Hui
    and Shutong Jiang and Zhaohai Li and Mingsheng Li and Mei Li
    and Kaixin Li and Zicheng Lin and Junyang Lin and Xuejing Liu
    and Jiawei Liu and Chenglong Liu and Yang Liu and Dayiheng Liu
    and Shixuan Liu and Dunjie Lu and Ruilin Luo and Chenxu Lv
    and Rui Men and Lingchen Meng and Xuancheng Ren and Xingzhang Ren
    and Sibo Song and Yuchong Sun and Jun Tang and Jianhong Tu
    and Jianqiang Wan and Peng Wang and Pengfei Wang and Qiuyue Wang
    and Yuxuan Wang and Tianbao Xie and Yiheng Xu and Haiyang Xu
    and Jin Xu and Zhibo Yang and Mingkun Yang and Jianxin Yang
    and An Yang and Bowen Yu and Fei Zhang and Hang Zhang
    and Xi Zhang and Bo Zheng and Humen Zhong and Jingren Zhou
    and Fan Zhou and Jing Zhou and Yuanzhi Zhu and Ke Zhu},
  journal = {arXiv preprint arXiv:2511.21631},
  year = {2025},
  doi = {10.48550/arXiv.2511.21631}
}

@article{fu2026videommev2,
  title = {{Video-MME-v2}: Towards the Next Stage in Benchmarks
           for Comprehensive Video Understanding},
  author = {Chaoyou Fu and Haozhi Yuan and Yuhao Dong
    and Yi-Fan Zhang and Yunhang Shen and Xiaoxing Hu
    and Xueying Li and Jinsen Su and Chengwu Long
    and Xiaoyao Xie and Yongkang Xie and Xiawu Zheng
    and Xue Yang and Haoyu Cao and Yunsheng Wu
    and Ziwei Liu and Xing Sun and Caifeng Shan and Ran He},
  journal = {arXiv preprint arXiv:2604.05015},
  year = {2026},
  doi = {10.48550/arXiv.2604.05015}
}

\clearpage
\appendix


\startcontents[appendix]

\section*{Appendix Contents}

\printcontents[appendix]{}{1}{%
    \setcounter{tocdepth}{2}%
}

\clearpage


\section{Extended Related Work}
\label{app:related_work}

\subsection{Post-Training for VideoLLMs}
\label{app:video_post_training}
RLVR supports VideoLLM post-training through outcome-level feedback on model-generated trajectories.
Video-R1 ~\citep{feng2025videor1} encourages temporal reasoning, VideoRFT ~\citep{wang2025videorft} rewards consistency between reasoning and visual evidence, and DeepVideo-R1 ~\citep{park2025deepvideor1} improves optimization through a regressive objective and difficulty-aware augmentation.
For localization, Time-R1 ~\citep{wang2025timer1} targets temporal grounding, while VideoChat-R1 ~\citep{li2025videochatr1} strengthens spatiotemporal perception through multi-task reinforcement fine-tuning.
Video-OPD ~\citep{li2026videoopd} provides dense teacher guidance for efficient temporal grounding, and VisionCoach ~\citep{lee2026visioncoach} combines visual prompting, grounding rewards, and self-distillation to improve evidence localization.

Open-o3 Video ~\citep{meng2025openo3video} unifies answer and spatiotemporal supervision through reasoning traces containing timestamps, objects, and bounding boxes, with rewards jointly evaluating answer correctness and grounding quality.
MeSD follows this paradigm using the same training data and annotations.
It organizes answer and grounding information into evidence-conditioned self-teacher contexts for token-level guidance along student-generated trajectories.


\subsection{On-Policy and Privileged-Context Self-Distillation}
\label{app:self_distillation}

On-policy distillation addresses the mismatch between distillation sequences and student-generated sequences at inference.
\textbf{Generalized Knowledge Distillation (GKD)} ~\citep{agarwal2024gkd} provides teacher-distribution supervision at student-generated prefixes.
\textbf{Video-OPD} ~\citep{li2026videoopd} applies this principle using a stronger external teacher.
\textbf{On-Policy Self-Distillation (OPSD)} ~\citep{zhao2026opsd} uses the same model under different contexts, conditioning the teacher on privileged information, such as reference solutions, to supervise student rollouts.
\textbf{Self-Distillation Policy Optimization (SDPO)} ~\citep{hubotter2026sdpo} converts textual environment feedback into dense supervision through a feedback-conditioned self-teacher.
Both derive teacher guidance from contextual information without a separately trained, stronger model.

Multimodal self-distillation extends this principle to visual evidence.
\textbf{Vision-OPD} ~\citep{yuan2026visionopd} transfers crop-conditioned predictions to a student processing full images.
\textbf{Video-OPSD} ~\citep{wang2026videoopsd} conditions its self-teacher on annotated evidence frames and weights distillation by token reliance on visual evidence.
\textbf{S³T} ~\citep{venkatraman2026s3t} uses denser temporal sampling to supervise a sparse-view student without labels or rewards.
\textbf{VISD} ~\citep{lin2026visd} incorporates structured feedback from a video-aware judge to diagnose answer consistency and spatiotemporal grounding.
These methods construct visual self-teachers through distinct contextual evidence.
MeSD organizes existing answer and grounding annotations into separate contexts to explicitly compare and fuse evidence-conditioned teacher preferences.


\subsection{Evidence Aggregation and Adaptive Supervision}
\label{app:multi_view_distillation}

Multi-teacher on-policy distillation combines complementary capabilities on student-generated trajectories.
\textbf{MOPD}~\citep{ma2026mopd} merges independently trained specialists into one student, while \textbf{MT-SDPO} ~\citep{he2026mtsdpo} conditions an EMA self-teacher on verified teacher feedback.
\textbf{AVSD} ~\citep{nguyen2026avsd} aggregates privileged views for self-distillation.
A single video-evidence context leaves source contributions and cross-evidence agreement, complementarity, and conflict implicit.
MeSD separates evidence-conditioned views and fuses temporal and spatial preference shifts relative to an answer-conditioned semantic anchor.

For adaptive supervision, \textbf{VG-OPD} ~\citep{xu2026vgopd} uses verification to select eligible experts and localize token-level supervision.
\textbf{RLSD} ~\citep{yang2026rlsd} and \textbf{VISD} ~\citep{lin2026visd} refine token credit while preserving reward-derived update directions; \textbf{H²SD} ~\citep{cai2026h2sd} explores outcome-dependent asymmetric self-distillation.
Failure feedback alone does not identify preferred alternatives.
Based on verification and evidence availability, MeSD uses its fused teacher distribution for reward-anchored credit refinement on successful and indeterminate trajectories, and reverse-KL correction only on eligible, verified failures.





\section{Implementation and Training Details}
\label{app:implementation}

\subsection{Training Data and Evidence Construction}
\label{app:data_evidence}

\begin{table}[htbp]
    \centering
    \caption{Task composition and annotation types of STGR-RL.
    Answer denotes the semantic answer used in the answer context.}
    \label{tab:training_data}

    \small
    \renewcommand{\arraystretch}{1.1}
    \setlength{\tabcolsep}{4pt}

    \begin{tabular}{@{}l r c c c@{}}
        \toprule
        \textbf{Task type} &
        \textbf{Samples} &
        \textbf{Answer} &
        \textbf{Temporal} &
        \textbf{Spatial} \\
        \midrule

        General video QA (MCQ)
        & 13,000 & Option & -- & -- \\

        General video QA (free-form)
        & 2,000 & Text & -- & -- \\

        Temporal QA
        & 2,280 & -- & Time intervals & -- \\

        Temporal QA (MCQ)
        & 2,904 & Option & Time intervals & -- \\

        Visual QA
        & 5,000 & Object name & -- & Object boxes \\

        Joint spatio-temporal QA
        & 12,047 & Text & Keyframe timestamps & Object boxes \\

        \midrule
        \textbf{Total}
        & \textbf{37,231} & & & \\
        \bottomrule
    \end{tabular}
\end{table}

We follow Open-o3-Video ~\citep{meng2025openo3video} and use
its STGR-RL training data, which covers video question answering
and temporal and spatial grounding.
For a fair comparison, we retain the same training data and
existing annotations without introducing any additional
human-written or model-generated annotations.

We construct the privileged teacher contexts directly from
the available annotations.
The Answer context contains the semantic answer, while the
Answer+Temporal and Answer+Spatial contexts additionally
include the available temporal annotations and object-level
bounding boxes, respectively.
Temporal evidence specifies relevant intervals or keyframe
timestamps, and spatial evidence preserves the associations
between annotated objects, bounding boxes, and their
corresponding frames.
Thus, MeSD reorganizes the existing supervision into
separate evidence-conditioned contexts without expanding
the annotation set.

Table~\ref{tab:training_data} summarizes the task composition
and annotation types of the training set.
It contains 37,231 samples, including 32,231 video-based
samples and 5,000 image-based spatial grounding samples.
Annotation availability varies across tasks: general video
QA provides semantic answers, temporal QA supplies time
intervals, and joint spatio-temporal QA includes keyframe
timestamps and object-level bounding boxes.
For pure temporal QA, the answer is itself a time interval
and therefore contributes temporal evidence rather than
a separate semantic answer.

\begin{table}[htbp]
    \centering
    \caption{Training configuration of MeSD.}
    \label{tab:training_hyperparameters}

    \small
    \renewcommand{\arraystretch}{1.18}
    \setlength{\tabcolsep}{5pt}

    \begin{tabularx}{\linewidth}{
        @{}X l
        @{\hspace{1.5em}}
        X l@{}
    }
        \toprule
        \multicolumn{2}{c}{\textbf{Training and rollout}} &
        \multicolumn{2}{c}{\textbf{Self-distillation}} \\
        \cmidrule(r{0.7em}){1-2}
        \cmidrule(l{0.7em}){3-4}

        \textbf{Parameter} & \textbf{Value} &
        \textbf{Parameter} & \textbf{Value} \\
        \midrule

        Training epochs & 1 &
        EMA update rate & 0.01 \\

        Optimizer & AdamW &
        Initial strength $\lambda_0$ & 0.5 \\

        Learning rate & $10^{-6}$ &
        Annealing steps & 1,000 \\

        LR schedule & Cosine &
        Token-weight clipping & $[0.8,\,1.2]$ \\

        Prompt batch size & 16 &
        Correction strength $\gamma$ & 0.8 \\

        Rollouts per prompt & 4 &
        Correction warm-up & 100 steps \\

        Sampling temperature & 1.0 &
        Teacher cutoff $H$ & 1,000 steps \\

        Video sampling & 2 FPS, $\leq 16$ frames &
        & \\

        Max.\ response length & 768 tokens &
        & \\

        \bottomrule
    \end{tabularx}
\end{table}

\subsection{Hyperparameters and Training Schedule}
\label{app:hyperparameters}

Table~\ref{tab:training_hyperparameters} summarizes the main
hyperparameters for our MeSD-7B experiments.
We use a staged training schedule that combines early
self-teacher guidance with subsequent reward-based optimization.
During the first 1,000 steps, trajectories eligible for
failure-conditioned correction are optimized with reverse-KL
distillation, while the remaining trajectories retain
reward-anchored credit refinement.

For credit refinement, the mixing coefficient $\lambda_s$
is initialized to $0.5$ and linearly annealed over the first
1,000 steps, gradually reducing the influence of teacher-based
token reweighting. The token weights are clipped to
$[0.8,\,1.2]$ to bound the modulation while preserving
the sign of the reward-derived advantage.
For eligible verified failures, the correction coefficient
$\gamma_s$ is linearly increased to $0.8$ over the first
100 steps and then held constant while teacher supervision
remains active. This warm-up gradually introduces
distribution-level correction during early training.

The shared self-teacher is updated by an exponential moving
average with an update rate of $0.01$, corresponding to
an EMA decay coefficient of $0.99$.
After step 1,000, both teacher-based credit refinement
and failure-conditioned distillation are disabled.
All subsequent updates use standard GRPO, completing
training under the original reward objective.


\subsection{Evidence-Conditioned Teacher Prompts}
\label{app:teacher_prompts}

We construct three teacher contexts: Answer,
Answer+Temporal, and Answer+Spatial.
All contexts retain the original task-specific system prompt,
visual input, and question, while appending different
privileged evidence to the user message.
The Answer context provides the verified semantic answer.
The Answer+Temporal context additionally includes annotated
temporal intervals or keyframe timestamps, whereas the
Answer+Spatial context adds object-level bounding boxes
associated with the corresponding image or annotated keyframe.
These boxes are expressed in the pixel coordinates of the
processed input image.

\begin{figure}[t]
    \centering
    \includegraphics[width=\textwidth]{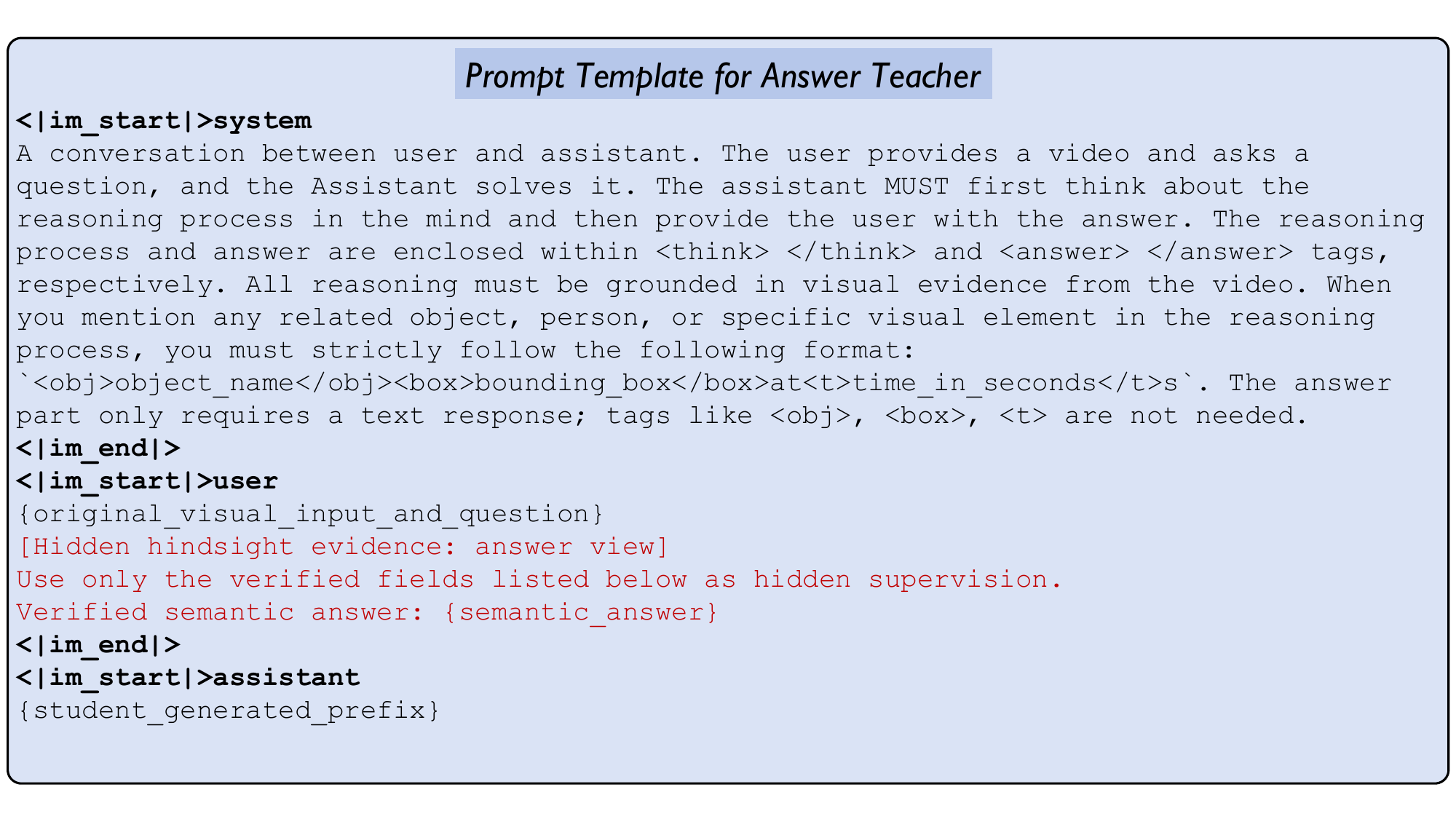}
    \caption{Answer teacher prompt with privileged semantic answers.}
    \label{fig:answer_teacher_prompt}
\end{figure}

\begin{figure}[t]
    \centering
    \includegraphics[width=\textwidth]{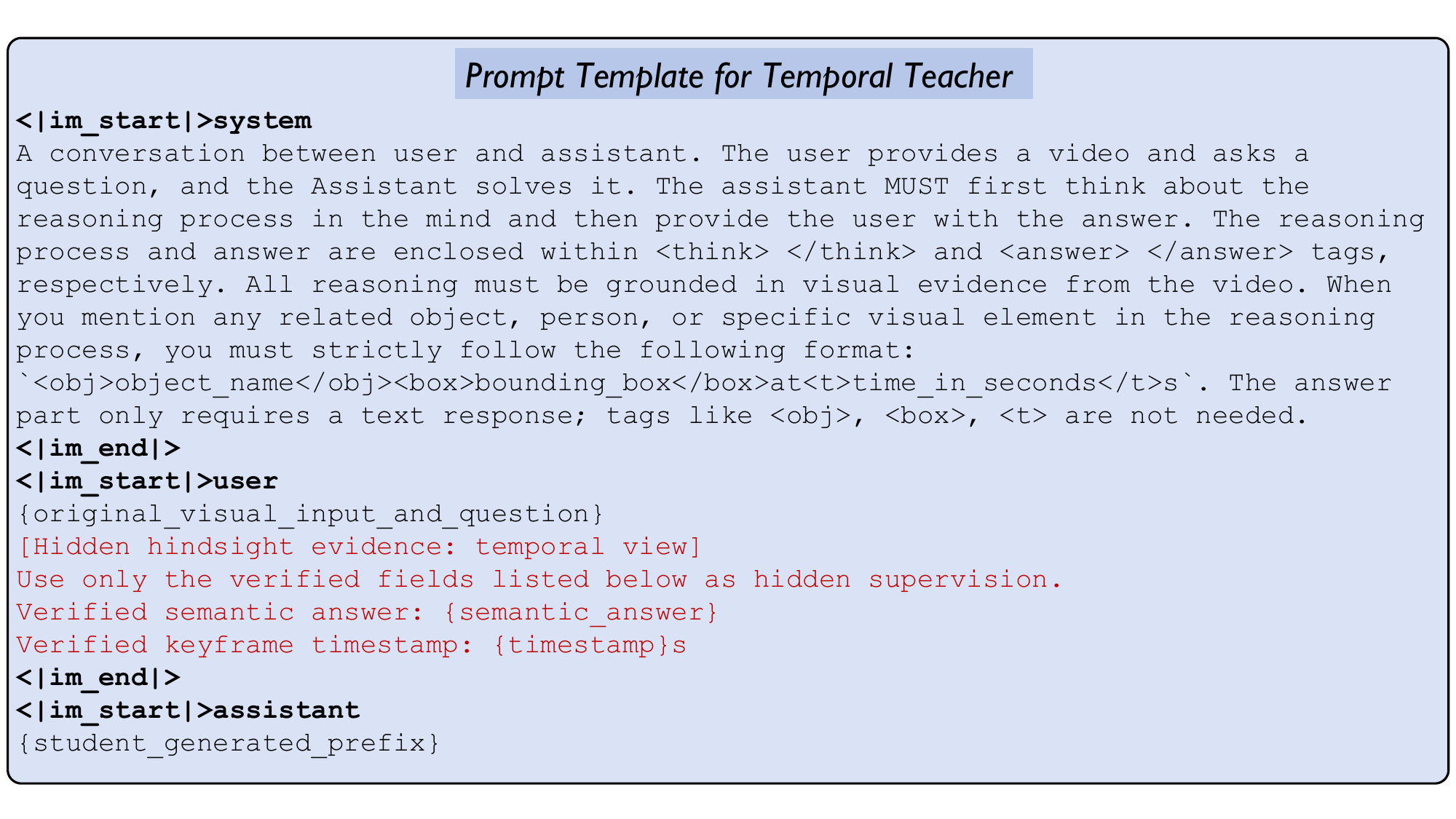}
\caption{Answer+Temporal teacher prompt with semantic answers
and keyframe timestamps.}
\label{fig:temporal_teacher_prompt}
\end{figure}

\begin{figure}[t]
    \centering
    \includegraphics[width=\textwidth]{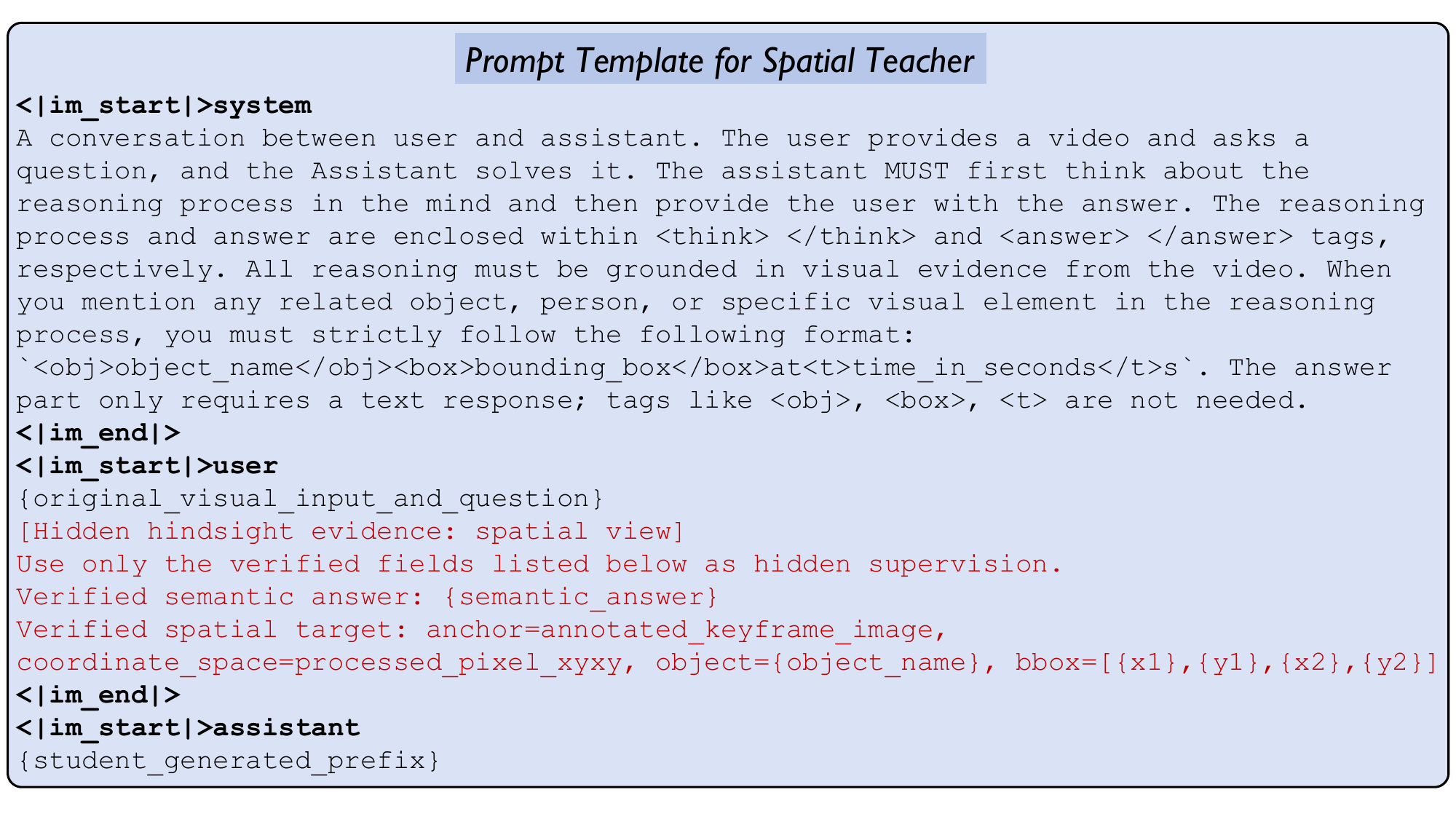}
\caption{Answer+Spatial teacher prompt with semantic answers
and object-level bounding boxes.}
\label{fig:spatial_teacher_prompt}
\end{figure}
The accompanying prompt templates illustrate the three
contexts for joint spatio-temporal question answering.
Braced placeholders are populated from existing annotations;
unavailable fields are omitted, and keyframe or spatial-target
entries are repeated when multiple annotations are available.
All three contexts are evaluated by the same EMA self-teacher
on the same student-generated prefix, without generating
additional teacher rollouts.
The privileged evidence is used only during training and
is not provided to the student at inference time.




\section{Supplementary Model Details}
\label{app:model_details}

This section provides additional details on teacher distribution computation, trajectory verification, and the two optimization branches, using the notation established in the main text.

\subsection{Teacher Distribution Computation}
\label{app:teacher_distribution}

At position $t$ in trajectory $i$, the student predicts the next token according to $\pi_\theta(\cdot\mid x_i,y_{i,<t})$. The evidence-conditioned teacher distributions $\bar q^k_{i,t}$, $k\in{A,T,S}$, are defined in the main text using their respective contexts $E_i^k$. These teachers share parameters $\bar\theta$ and predict the next token conditioned on the same student-generated prefix $y_{i,<t}$. Below, the superscripts $\mathrm{cr}$ and $\mathrm{rkl}$ distinguish the probability representations used for credit refinement and failure-conditioned distillation, respectively.

\noindent\textbf{Shared candidate support.}
Let $\mathcal{K}_i\subseteq\{A,T,S\}$ denote the available
teacher views for trajectory $i$.
For each optimization branch
$b\in\{\mathrm{cr},\mathrm{rkl}\}$, we construct
\begin{equation}
\begin{aligned}
\mathcal{U}_{i,t}^{b}
={}&
\{y_{i,t}\}
\cup
\operatorname{TopK}_{K_b}
\bigl(\pi_\theta(\cdot\mid x_i,y_{i,<t})\bigr)
&\cup
\bigcup_{k\in\mathcal{K}_i}
\operatorname{TopK}_{K_b}
\bigl(\bar q^k_{i,t}\bigr),
\end{aligned}
\label{eq:app_candidate_support}
\end{equation}
where $\operatorname{TopK}_{K_b}$ returns the token IDs
with the $K_b$ highest probabilities in a distribution.
The union removes duplicate IDs, excludes unavailable views,
and always includes the sampled token $y_{i,t}$.
We set $K_{\mathrm{cr}}=16$ and $K_{\mathrm{rkl}}=32$;
these values specify the number of top-ranked candidates
selected from each distribution, not the size of the final
candidate set.
Candidate selection is performed without gradient tracking.

\noindent\textbf{Probabilities for credit refinement.}
We renormalize the student and available teacher distributions
over the shared candidate set $\mathcal{U}_{i,t}^{\mathrm{cr}}$:
\begin{equation}
\begin{aligned}
\bar p_{i,t}^{\mathrm{cr}}(z)
&=
\frac{\pi_\theta(z\mid x_i,y_{i,<t})}
{\sum_{u\in\mathcal{U}_{i,t}^{\mathrm{cr}}}
 \pi_\theta(u\mid x_i,y_{i,<t})},
\\
\bar q_{i,t}^{k,\mathrm{cr}}(z)
&=
\frac{\bar q_{i,t}^{k}(z)}
{\sum_{u\in\mathcal{U}_{i,t}^{\mathrm{cr}}}
 \bar q_{i,t}^{k}(u)},
\qquad k\in\mathcal{K}_i,
\end{aligned}
\label{eq:app_cr_normalization}
\end{equation}
for $z\in\mathcal{U}_{i,t}^{\mathrm{cr}}$.
We apply the teacher distribution fusion described in the
main text to $\bar q_{i,t}^{k,\mathrm{cr}}$, obtaining
$\bar q_{i,t}^{\mathrm{mv},\mathrm{cr}}$.
Accordingly, all fusion quantities,
$d^k(z)$, $G(z)$, $D(z)$, $C(z)$, $H(z)$, and $f(z)$,
are computed from these candidate-normalized teacher
distributions.

The fused teacher distribution
$\bar q_{i,t}^{\mathrm{mv},\mathrm{cr}}$
and the student distribution $\bar p_{i,t}^{\mathrm{cr}}$
define the log-probability difference $\Delta_{i,t}$
used in the main text.
This difference is detached before computing the token
weight $w_{i,t}$ and the refined advantage
$A_{i,t}^{\mathrm{cr}}$.
The policy ratio $\rho_{i,t}$, however, uses the
full-vocabulary probabilities of the sampled token under
the current and rollout policies, rather than their
candidate-normalized probabilities.

\noindent\textbf{Probabilities for failure-conditioned distillation.}
We define the coarse probability space
$\Omega_{i,t}
=\mathcal{U}_{i,t}^{\mathrm{rkl}}\cup\{\bot\}$,
where the tail category $\bot$ groups all tokens outside
the candidate set.
The student distribution on this space is
\begin{equation}
\bar p_{i,t}^{\mathrm{rkl}}(z)
=
\begin{cases}
\pi_\theta(z\mid x_i,y_{i,<t}),
& z\in\mathcal{U}_{i,t}^{\mathrm{rkl}},
\\
1-\displaystyle\sum_{u\in\mathcal{U}_{i,t}^{\mathrm{rkl}}}
\pi_\theta(u\mid x_i,y_{i,<t}),
& z=\bot.
\end{cases}
\label{eq:app_tail_distribution}
\end{equation}
Each teacher distribution
$\bar q_{i,t}^{k,\mathrm{rkl}}$ is constructed in the same
way from $\bar q_{i,t}^{k}$.
This representation preserves the full-vocabulary probability
of each explicit candidate and assigns the remaining
probability mass to $\bot$, rather than renormalizing
over the candidate set alone.

We apply teacher distribution fusion to
$\bar q_{i,t}^{k,\mathrm{rkl}}$ over $\Omega_{i,t}$,
including the tail category, to obtain
$\bar q_{i,t}^{\mathrm{mv},\mathrm{rkl}}$.
We then apply a probability floor and renormalize the fused
distribution to obtain the target used in the main-text
reverse-KL objective:
\begin{equation}
\widehat q_{i,t}^{\mathrm{rkl}}(z)
=
\frac{
\max\{
\bar q_{i,t}^{\mathrm{mv},\mathrm{rkl}}(z),
\epsilon_{\mathrm{tar}}
\}
}{
\sum_{u\in\Omega_{i,t}}
\max\{
\bar q_{i,t}^{\mathrm{mv},\mathrm{rkl}}(u),
\epsilon_{\mathrm{tar}}
\}
},
\label{eq:app_target_floor}
\end{equation}
where $\epsilon_{\mathrm{tar}}=10^{-8}$.
The target $\widehat q_{i,t}^{\mathrm{rkl}}$ is detached,
while gradients flow through the student distribution
$\bar p_{i,t}^{\mathrm{rkl}}$, including its tail mass.
The probability floor is applied only to the teacher target.
Thus, the two optimization branches use the same fusion
rule but construct their fused targets separately,
using their respective candidate sets and probability
representations.

\subsection{Trajectory Verification}
\label{app:trajectory_verification}

We distinguish three decisions: outcome verification
assigns a trajectory label, data validation determines
whether the training example is usable, and correction
selection determines whether to apply failure-conditioned
distillation.

\noindent\textbf{Outcome verification.}
Unlike sequence-level rewards, which provide graded
feedback for policy optimization, verification assigns
Success, Failure, or Indeterminate.
For supported tasks, a trajectory is labeled Success
if all required components pass, Failure if any component
explicitly fails, and Indeterminate otherwise.
Table~\ref{tab:verification_rules} summarizes the
task-specific criteria and outcome-based eligibility
for reverse-KL (RKL) distillation.

\begin{table*}[htbp]
\centering
\small
\setlength{\tabcolsep}{6pt}
\renewcommand{\arraystretch}{1.15}
\caption{Task-specific outcome verification and RKL eligibility
under the strict setting.}
\label{tab:verification_rules}

\begin{tabularx}{\textwidth}{
    @{}
    >{\raggedright\arraybackslash}p{0.19\textwidth}
    >{\raggedright\arraybackslash}X
    >{\raggedright\arraybackslash}X
    >{\raggedright\arraybackslash}X
    @{}
}
\toprule
\textbf{Task}
& \textbf{Success}
& \textbf{Failure}
& \textbf{RKL eligibility} \\
\midrule

Video MCQ
& Correct option
& Incorrect option
& Incorrect option \\

\addlinespace
Temporal localization
& $\mathrm{tIoU}\geq 0.5$
& $\mathrm{tIoU}=0$
& $\mathrm{tIoU}=0$ \\

\addlinespace
Temporal MCQ
& Correct option \textbf{and}\newline
  $\mathrm{tIoU}\geq 0.5$
& Incorrect option \textbf{or}\newline
  $\mathrm{tIoU}=0$
& Incorrect option \textbf{and}\newline
  $\mathrm{tIoU}=0$ \\

\midrule

Spatial localization
& Not assigned
& Not assigned
& Not eligible \\

\addlinespace
Other composite or free-form tasks
& Not assigned
& Not assigned
& Not eligible \\

\bottomrule
\end{tabularx}
\end{table*}

The verifier requires exactly one nonempty
\texttt{<answer>} block.
Multiple-choice outputs must contain a parseable option
from A--H.
Temporal outputs must contain exactly one interval
in the expected timestamp format, with finite endpoints
and an end time strictly greater than the start time.
For temporal multiple-choice QA, the option and interval
are extracted separately from the \texttt{Correct Option:}
field and the answer block.
An unparseable component satisfies neither the success
nor the failure criterion; another explicitly incorrect
component can still establish Failure.
Missing or malformed intervals are never treated as
zero-overlap predictions.
Tasks without explicit verification rules retain their
task-specific rewards for policy optimization.

\noindent\textbf{Data validity.}
Malformed model outputs are distinct from invalid
training data.
The former may remain valid trajectories with uncertain
outcomes, whereas unparseable ground-truth options,
invalid ground-truth intervals, and visual loading failures
are flagged as invalid.
Visual loading failures additionally receive a skip flag.
Any Indeterminate label attached to an invalid row is
a placeholder, not an uncertain prediction on valid data.

Required reward metadata is also validated.
Spatial localization requires annotation and processed
image sizes.
Temporal--spatial free-form QA requires frame sizes,
nonempty keyframe annotations, an object annotation mapping,
and training progress for reward scheduling.
Under strict validation, missing or malformed required
metadata raises an error rather than silently yielding
a zero reward.

\noindent\textbf{Correction selection.}
A Failure label alone does not guarantee RKL eligibility.
In particular, temporal multiple-choice QA requires
both an incorrect option and a valid interval with
zero tIoU under the strict setting.
Responses with only one incorrect component are therefore
labeled Failure but excluded from correction.

Let $v_i$ indicate data validity, $e_i$ denote outcome-based
eligibility from Table~\ref{tab:verification_rules},
and $h_i$ indicate whether the teacher evidence required
to construct a correction target is available.
During teacher supervision, the correction indicator is
\begin{equation}
g_i=v_i e_i h_i.
\label{eq:app_correction_selection}
\end{equation}
Evidence availability is checked independently of the
verification outcome.

For valid trajectories, $g_i=1$ selects
failure-conditioned distillation in place of the policy
loss, whereas $g_i=0$ retains token-level advantage
refinement.
This fallback includes Success and Indeterminate
trajectories, failures excluded by the strict eligibility
rule, and otherwise eligible failures lacking the required
teacher evidence.
If no usable teacher signal is available for refinement,
the original reward-derived advantage is retained.



%
%
%
%

\subsection{Analysis of Credit Refinement and Failure Correction}
\label{app:optimization_analysis}

We examine the complementary roles of the refined advantage
$A_{i,t}^{\mathrm{cr}}$ and the failure-conditioned loss
$\ell_i^{\mathrm{rkl}}$.
Credit refinement adjusts reward-derived token credit,
whereas failure correction provides a distribution-level
learning signal for eligible verified failures.

\noindent\textbf{Sign preservation and bounded refinement.}
The clipped token weight lies in
$[1-\epsilon_w,1+\epsilon_w]$.
Thus, for $A_i\neq 0$, the refined advantage satisfies
\begin{equation}
1-\lambda_s\epsilon_w
\leq
\frac{A_{i,t}^{\mathrm{cr}}}{A_i}
\leq
1+\lambda_s\epsilon_w.
\label{eq:app_refinement_bound}
\end{equation}
For $\lambda_s\in[0,1]$ and $\epsilon_w\in[0,1)$,
the lower bound is strictly positive, yielding
\begin{equation}
\begin{aligned}
\operatorname{sign}(A_{i,t}^{\mathrm{cr}})
&=\operatorname{sign}(A_i),\\
\left|A_{i,t}^{\mathrm{cr}}-A_i\right|
&\leq\lambda_s\epsilon_w|A_i|.
\end{aligned}
\label{eq:app_advantage_properties}
\end{equation}
When $A_i=0$, refinement gives
$A_{i,t}^{\mathrm{cr}}=0$, so both properties in
Eq.~\eqref{eq:app_advantage_properties} still hold.
These results establish sign preservation and a bounded
change in the advantage coefficients; they do not imply
that the aggregate parameter-gradient direction is unchanged.
With the initial settings $\lambda_s=0.5$ and
$\epsilon_w=0.2$, the ratio
$A_{i,t}^{\mathrm{cr}}/A_i$ lies in $[0.9,1.1]$ for
nonzero $A_i$.
This interval contracts to $\{1\}$ as $\lambda_s$
approaches zero.
The teacher-derived refinement factors are detached
throughout optimization.

\noindent\textbf{Correction beyond multiplicative refinement.}
Identical rewards within a rollout group yield zero
group-relative advantages.
Multiplicative refinement then gives
$A_{i,t}^{\mathrm{cr}}=0$ regardless of $\Delta_{i,t}$,
so the corresponding credit-refined policy-loss terms
provide no gradient signal.
In contrast, $\ell_i^{\mathrm{rkl}}$ is independent of $A_i$.

To characterize this distinction, write the per-token
reverse-KL loss as
\begin{equation}
\begin{aligned}
\ell_{i,t}^{\mathrm{rkl}}
&=
D_{\mathrm{KL}}
\left(
\bar p_{i,t}^{\mathrm{rkl}}
\,\middle\|\,
\operatorname{sg}
[\widehat q_{i,t}^{\mathrm{rkl}}]
\right),
\\
\ell_i^{\mathrm{rkl}}
&=
\frac{1}{T_i}
\sum_{t=1}^{T_i}\ell_{i,t}^{\mathrm{rkl}}.
\end{aligned}
\label{eq:app_token_rkl}
\end{equation}
For this analysis, hold the coarse alphabet $\Omega_{i,t}$
and the detached teacher target fixed.
Parameterize a strictly positive coarse student distribution
using auxiliary logits $\eta_{i,t}(z)$:
\begin{equation}
\bar p_{i,t}^{\mathrm{rkl}}(z)
=
\frac{\exp(\eta_{i,t}(z))}
{\sum_{u\in\Omega_{i,t}}\exp(\eta_{i,t}(u))}.
\label{eq:app_coarse_parameterization}
\end{equation}
Differentiating with respect to these logits gives
\begin{equation}
\frac{\partial\ell_{i,t}^{\mathrm{rkl}}}
{\partial\eta_{i,t}(z)}
=
\bar p_{i,t}^{\mathrm{rkl}}(z)
\left[
\log
\frac{\bar p_{i,t}^{\mathrm{rkl}}(z)}
{\widehat q_{i,t}^{\mathrm{rkl}}(z)}
-
\ell_{i,t}^{\mathrm{rkl}}
\right].
\label{eq:app_rkl_gradient}
\end{equation}
This gradient does not depend on $A_i$.
Since all student probabilities are positive,
the gradient vanishes in every coordinate if and only if
the log-probability ratio is constant across $\Omega_{i,t}$.
Normalization then implies
$\bar p_{i,t}^{\mathrm{rkl}}
=\widehat q_{i,t}^{\mathrm{rkl}}$.
Thus, for eligible verified failures, distribution mismatch
can supply a correction signal even when credit refinement
provides none.
This conclusion concerns the distribution-level objective;
it does not guarantee a correct answer or a nonzero
aggregate gradient with respect to the model parameters.

\noindent\textbf{Selective correction and return to GRPO.}
For valid trajectories, $g_i$ selects the optimization branch:
$g_i=1$ replaces $\ell_i^{\mathrm{cr}}$ with
$\gamma_s\ell_i^{\mathrm{rkl}}$, whereas $g_i=0$
retains credit refinement.
The latter includes failures excluded by the strict rules
in Section~\ref{app:trajectory_verification}.
When no usable teacher signal is available,
$\Delta_{i,t}=0$, so $w_{i,t}=1$ and
$A_{i,t}^{\mathrm{cr}}=A_i$.
After teacher supervision is disabled, setting
$\lambda_s=0$ and $g_i=0$ for all trajectories
reduces $\mathcal{L}_s(\theta)$ to standard GRPO.


\section{Evaluation Details}
\label{app:evaluation}

Our evaluation examines whether multi-evidence
self-distillation improves answer correctness and
spatio-temporal localization, and whether these gains
extend to broader video understanding tasks.
We describe the benchmarks, comparison methods, and
evaluation metrics below.

\subsection{Benchmarks}
\label{app:benchmarks}

\noindent\textbf{Joint spatiotemporal reasoning.}
V-STaR ~\citep{cheng2026vstar} jointly evaluates question answering and the localization of supporting temporal and spatial evidence.
It contains two reasoning chains: what--when--where and what--where--when.
We report results for both chains to assess MeSD under different evidence prediction orders and distinguish gains in answer accuracy from those in temporal and spatial grounding.

\noindent\textbf{Temporal grounding.}
We use TVGBench ~\citep{wang2025timer1} and Charades-STA ~\citep{gao2017tall} to evaluate temporal localization beyond joint reasoning.
TVGBench contains temporal grounding examples from multiple source datasets, while Charades-STA requires identifying the video interval described by a natural-language query.
These benchmarks complement V-STaR by directly assessing how well predicted intervals align with annotated events.

\noindent\textbf{General and long-video understanding.}
We use WorldSense ~\citep{hong2026worldsense}, VideoMMMU ~\citep{hu2025videommmu}, and LongVideo-Reason-eval (LRR) ~\citep{chen2025scalingrl} to assess whether evidence-based training also improves video question answering.
WorldSense covers recognition, understanding, and reasoning.
VideoMMMU evaluates knowledge acquisition from professional videos through perception, comprehension, and adaptation, while LRR tests reasoning over long videos.
We also evaluate on Video-MME-v2 ~\citep{fu2026videommev2}, which uses groups of related questions to assess capability consistency and reasoning coherence.
Together, these benchmarks assess whether MeSD's benefits extend beyond explicit grounding.

\subsection{Baselines}
\label{app:baselines}

\noindent\textbf{General and video-specialized models.}
We include Qwen2.5-VL-7B ~\citep{bai2025qwen25vl}, InternVL-2.5-8B ~\citep{chen2024internvl25}, and Oryx-1.5-7B ~\citep{liu2025oryx}, alongside video-specialized models such as VideoChat2 ~\citep{li2024mvbench}, Video-LLaMA3 ~\citep{zhang2025videollama3}, LLaVA-Video ~\citep{zhang2025llavavideo}, and TRACE ~\citep{guo2025trace}.
We also report results for GPT-4o ~\citep{openai2024gpt4o} and Gemini-2-Flash ~\citep{google2025gemini2flash} as proprietary references where available.
For Charades-STA, we include temporal grounding methods such as TimeChat ~\citep{ren2024timechat}, HawkEye ~\citep{wang2024hawkeye}, VTG-LLM ~\citep{guo2025vtgllm}, and VTimeLLM ~\citep{huang2024vtimellm}.
These comparisons position MeSD relative to a broad range of video understanding and localization methods.


\noindent\textbf{Video post-training and self-distillation.}
Our main post-training baselines include VideoRFT ~\citep{wang2025videorft}, Video-R1 ~\citep{feng2025videor1}, Open-o3-Video ~\citep{meng2025openo3video}, VisionCoach ~\citep{lee2026visioncoach}, and VISD ~\citep{lin2026visd}.
Our training setup builds on Open-o3-Video's grounded video reasoning framework.
VISD serves as a closely related self-distillation baseline, using teacher-derived signals to refine token-level credit while preserving reward-derived update directions.
Comparing MeSD with VISD assesses the added value of separating evidence-conditioned teacher preferences and introducing distribution-level correction for eligible, verified failures.

\noindent\textbf{Optimization and component comparisons.}
We also include GRPO and GSPO as reward-based optimization baselines.
Component ablations individually assess evidence separation, temporal and spatial supervision, gated preference aggregation, and failure-conditioned distillation.
These comparisons complement the model-level results by identifying the contributions of individual MeSD components.


\subsection{Evaluation Metrics}
\label{app:metrics}

\begingroup
\setlength{\abovedisplayskip}{5pt plus 1pt minus 1pt}
\setlength{\belowdisplayskip}{5pt plus 1pt minus 1pt}
\setlength{\abovedisplayshortskip}{3pt plus 1pt minus 1pt}
\setlength{\belowdisplayshortskip}{3pt plus 1pt minus 1pt}
\setlength{\jot}{2pt}

\noindent\textbf{Answer accuracy.}
We report answer accuracy for WorldSense, VideoMMMU,
and LRR, together with the category-level results
presented in the main tables.
V-STaR answer correctness is assessed using the
Qwen2.5-72B-Instruct evaluator in our evaluation pipeline.
Answer accuracy measures whether the final response is
correct, while the grounding metrics below assess
the predicted evidence separately.

\noindent\textbf{Temporal and spatial overlap.}
For a predicted interval $I_p$ and a reference interval
$I_g$, temporal intersection over union is defined as
\begin{equation}
    \operatorname{tIoU}(I_p,I_g)
    = \frac{|I_p \cap I_g|}{|I_p \cup I_g|}.
    \label{eq:eval_tiou}
\end{equation}
We report mean IoU on TVGBench and Charades-STA.
For Charades-STA, we additionally report
$R@0.3$, $R@0.5$, and $R@0.7$, measuring the fraction
of queries whose predicted interval reaches the
corresponding IoU threshold.
Spatial grounding on V-STaR is measured using bounding-box
IoU with the matching and aggregation rules of its
evaluation pipeline.

\noindent\textbf{Joint reasoning scores.}
For V-STaR, let $\mathcal{C}$ denote the set of evaluated
reasoning chains and $\mathcal{D}$ the set of evaluation
dimensions: answer accuracy, temporal IoU, and spatial IoU.
Let $m_{c,d}$ denote the component score for dimension
$d \in \mathcal{D}$ in chain $c \in \mathcal{C}$,
normalized to $[0,1]$.
Answer accuracy is shared across chains, whereas temporal
and spatial IoU are chain-specific.
We compute
\begin{align}
    \operatorname{mAM}
    &= \frac{1}{|\mathcal{C}|\,|\mathcal{D}|}
       \sum_{c \in \mathcal{C}}
       \sum_{d \in \mathcal{D}} m_{c,d},
    \label{eq:eval_mam}\\
    \operatorname{mLGM}
    &= -\frac{1}{|\mathcal{C}|\,|\mathcal{D}|}
       \sum_{c \in \mathcal{C}}
       \sum_{d \in \mathcal{D}}
       \log(1-m_{c,d}).
    \label{eq:eval_mlgm}
\end{align}
We report the individual dimensions alongside these
aggregates to distinguish answer correctness from
temporal and spatial grounding.
Both aggregates are multiplied by 100 for display.
mLGM is a logarithmic score rather than a percentage
accuracy and is not bounded by 100.

\noindent\textbf{Grouped video understanding scores.}
For Video-MME-v2, we report the grouped non-linear
score, Level 1--3 scores, capability consistency,
reasoning coherence, and average question-level accuracy.
The grouped scores account for correctness across related
questions and their dependency structure, providing
information beyond isolated answer accuracy.

\noindent\textbf{Cross-benchmark average.}
The average reported for WorldSense, VideoMMMU, LRR,
and TVGBench is the unweighted arithmetic mean of
WorldSense overall accuracy, VideoMMMU overall accuracy,
LRR accuracy, and TVGBench mIoU.
All four scores use the 0--100 scale.
Category-level scores are excluded from this average
to avoid counting the same benchmark multiple times.

\raggedbottom
\setcounter{section}{5}
\section{Additional Experimental Results}
\label{app:additional_results}

\subsection{Video-MME-v2 Results}
\label{app:video_mme_v2_results}

\begin{table*}[!htbp]
\setlength{\belowcaptionskip}{5pt}
\centering
\caption{Video-MME-v2 results with 64 sampled frames and no subtitles. Bold and underlining mark the best and second-best scores; $\Delta$ gives the gain over Qwen2.5-VL-7B.}
\label{tab:video_mme_v2}

\small
\renewcommand{\arraystretch}{1.15}
\setlength{\tabcolsep}{3pt}

\begin{NiceTabularX}{\textwidth}{
  @{}X[l]*{7}{c}@{}
}
\CodeBefore
    \rectanglecolor{blue!10}{7-1}{7-8}
    \rectanglecolor{gray!15}{8-1}{8-8}
\Body
\toprule
\textbf{Model}
& \textbf{Non-Lin Score}
& \textbf{Level 1}
& \textbf{Level 2}
& \textbf{Level 3}
& \textbf{Consistency}
& \textbf{Coherence}
& \textbf{Avg Acc} \\
\midrule

Qwen2.5-VL-7B  ~\citep{bai2025qwen25vl}
& 7.9
& 11.4
& 6.7
& 6.7
& 8.2
& 7.3
& 21.3 \\

VideoRFT-7B  ~\citep{wang2025videorft}
& 8.2
& 11.8
& 7.4
& 6.8
& 8.8
& 7.2
& 21.7 \\

Video-R1-7B  ~\citep{feng2025videor1}
& 7.7
& 11.1
& 6.7
& 6.4
& 8.0
& 7.2
& 20.7 \\

Open-o3-Video-7B$^{*}$ ~\citep{meng2025openo3video}
& 7.8
& 11.0
& 6.9
& 6.6
& 8.3
& 6.9
& 21.2 \\

VISD-7B ~\citep{lin2026visd}
& \underline{9.4}
& \underline{13.4}
& \underline{8.2}
& \underline{7.9}
& \underline{10.2}
& \underline{8.0}
& \underline{23.8} \\

\midrule

\textbf{MeSD-7B (Ours)}
& \textbf{12.7}
& \textbf{18.7}
& \textbf{10.0}
& \textbf{11.2}
& \textbf{12.2}
& \textbf{13.7}
& \textbf{28.4} \\

$\Delta$ vs. Qwen-VL
& +4.8
& +7.3
& +3.3
& +4.5
& +4.0
& +6.4
& +7.1 \\

\bottomrule
\end{NiceTabularX}
\end{table*}
\FloatBarrier

On Video-MME-v2, MeSD reaches 28.4 average accuracy, exceeding VISD
by 4.6 points and Qwen2.5-VL-7B by 7.1 points. Its gains also extend
to the non-linear score and each reported component, indicating stronger
performance under the grouped-question evaluation.

\subsection{Qwen3-VL Detailed Results}
\label{app:qwen3_detailed_results}

Tables~\ref{tab:qwen3_vstar} and~\ref{tab:qwen3_general} compare
each MeSD model with its base Qwen3-VL counterpart at 4B, 8B,
and 32B parameters. They report both spatiotemporal grounding
and the four-benchmark average.

\begin{table}[H]
\setlength{\belowcaptionskip}{5pt}
\centering
\caption{V-STAR results across Qwen3-VL model sizes. Chain1 follows what--when--where and Chain2 follows what--where--when. Bold marks the better score within each base/MeSD pair; $\Delta$ is the MeSD--base point difference.}
\label{tab:qwen3_vstar}
\small
\renewcommand{\arraystretch}{1.15}
\setlength{\tabcolsep}{4pt}
\begin{adjustbox}{max width=\linewidth}
\begin{NiceTabular}{@{}lccccccc@{}}
\CodeBefore
\rowcolor{blue!10}{4,7,10}
\rowcolor{gray!15}{5,8,11}
\Body
\toprule
\multirow{2}{*}[-2.7pt]{\textbf{Model}}
& \textbf{What}
& \multicolumn{2}{c}{\textbf{When (temporal IoU)}}
& \multicolumn{2}{c}{\textbf{Where (spatial IoU)}}
& \multicolumn{2}{c}{\textbf{Overall}} \\
\cmidrule(lr){3-4}
\cmidrule(lr){5-6}
\cmidrule(lr){7-8}
& \textbf{Acc.}
& \textbf{Chain1} & \textbf{Chain2}
& \textbf{Chain1} & \textbf{Chain2}
& \textbf{mAM} & \textbf{mLGM} \\
\midrule
Qwen3-VL-4B ~\citep{bai2025qwen3vl}
& 22.7
& 16.8 & 16.2
& 15.9 & 2.8
& 16.2 & 18.0 \\
\textbf{Qwen3-VL-MeSD-4B}
& \textbf{62.1}
& \textbf{23.1} & \textbf{23.8}
& \textbf{22.5} & \textbf{6.4}
& \textbf{33.3} & \textbf{46.6} \\
$\Delta$ vs. Qwen3-VL-4B
& $+39.4$
& $+6.3$ & $+7.6$
& $+6.6$ & $+3.6$
& $+17.2$ & $+28.6$ \\
\midrule
Qwen3-VL-8B ~\citep{bai2025qwen3vl}
& 31.9
& 21.7 & 21.1
& 5.5 & 0.7
& 18.8 & 21.9 \\
\textbf{Qwen3-VL-MeSD-8B}
& \textbf{60.0}
& \textbf{25.5} & \textbf{25.9}
& \textbf{25.4} & \textbf{6.3}
& \textbf{33.8} & \textbf{46.4} \\
$\Delta$ vs. Qwen3-VL-8B
& $+28.1$
& $+3.8$ & $+4.8$
& $+19.9$ & $+5.6$
& $+15.0$ & $+24.5$ \\
\midrule
Qwen3-VL-32B ~\citep{bai2025qwen3vl}
& 47.5
& 24.5 & 25.7
& 30.9 & 7.5
& 30.6 & 38.6 \\
\textbf{Qwen3-VL-MeSD-32B}
& \textbf{66.0}
& \textbf{29.2} & \textbf{29.4}
& \textbf{31.1} & \textbf{11.4}
& \textbf{38.9} & \textbf{55.8} \\
$\Delta$ vs. Qwen3-VL-32B
& $+18.5$
& $+4.7$ & $+3.7$
& $+0.2$ & $+3.9$
& $+8.3$ & $+17.2$ \\
\bottomrule
\end{NiceTabular}
\end{adjustbox}
\end{table}

\begin{table}[H]
\setlength{\belowcaptionskip}{5pt}
\centering
\caption{Results across Qwen3-VL model sizes on WorldSense, VideoMMMU, LRR, and TVGBench. Bold marks the better score within each pair, including ties; $\Delta$ is the MeSD--base difference. Avg and $\Delta$ are computed before rounding.}
\label{tab:qwen3_general}
\small
\renewcommand{\arraystretch}{1.15}
\setlength{\tabcolsep}{3pt}
\begin{adjustbox}{max width=\linewidth}
\begin{NiceTabular}{@{}l*{11}{c}@{}}
\CodeBefore
\rowcolor{blue!10}{4,7,10}
\rowcolor{gray!15}{5,8,11}
\Body
\toprule
\multirow{2}{*}[-2.7pt]{\textbf{Model}}
& \multicolumn{4}{c}{\textbf{WorldSense}}
& \multicolumn{4}{c}{\textbf{VideoMMMU}}
& \textbf{LRR}
& \textbf{TVGBench}
& \multirow{2}{*}[-2.7pt]{\textbf{Avg}} \\
\cmidrule(lr){2-5}
\cmidrule(lr){6-9}
\cmidrule(lr){10-10}
\cmidrule(lr){11-11}
& \textbf{Overall} & Recog. & Underst. & Reason.
& \textbf{Overall} & Percep. & Adapt. & Compreh.
& Acc. & mIoU & \\
\midrule
Qwen3-VL-4B ~\citep{bai2025qwen3vl}
& 39.7 & 39.4 & 40.9 & 38.9
& 55.2 & 76.7 & 38.7 & 50.3
& 69.3 & 13.7 & 44.5 \\
\textbf{Qwen3-VL-MeSD-4B}
& \textbf{41.8} & \textbf{42.7} & \textbf{41.0} & \textbf{41.1}
& \textbf{59.7} & \textbf{78.7} & \textbf{45.3} & \textbf{55.0}
& \textbf{75.2} & \textbf{29.9} & \textbf{51.7} \\
$\Delta$ vs. Qwen3-VL-4B
& $+2.1$ & $+3.3$ & $+0.1$ & $+2.2$
& $+4.5$ & $+2.0$ & $+6.6$ & $+4.7$
& $+5.9$ & $+16.2$ & $+7.2$ \\
\midrule
Qwen3-VL-8B ~\citep{bai2025qwen3vl}
& 41.4 & 40.1 & 45.5 & 38.9
& 61.3 & 76.7 & \textbf{48.7} & 58.7
& 70.2 & 17.1 & 47.5 \\
\textbf{Qwen3-VL-MeSD-8B}
& \textbf{44.1} & \textbf{42.8} & \textbf{48.1} & \textbf{41.9}
& \textbf{62.3} & \textbf{79.0} & 47.7 & \textbf{60.3}
& \textbf{76.1} & \textbf{30.2} & \textbf{53.2} \\
$\Delta$ vs. Qwen3-VL-8B
& $+2.7$ & $+2.7$ & $+2.6$ & $+3.0$
& $+1.0$ & $+2.3$ & $-1.0$ & $+1.6$
& $+5.9$ & $+13.1$ & $+5.7$ \\
\midrule
Qwen3-VL-32B ~\citep{bai2025qwen3vl}
& 44.6 & 44.4 & \textbf{48.0} & \textbf{41.4}
& 65.2 & 84.0 & 51.3 & 60.3
& 74.4 & 31.2 & 53.9 \\
\textbf{Qwen3-VL-MeSD-32B}
& \textbf{46.1} & \textbf{47.7} & 47.9 & \textbf{41.4}
& \textbf{72.8} & \textbf{85.7} & \textbf{58.7} & \textbf{74.0}
& \textbf{79.4} & \textbf{39.7} & \textbf{59.5} \\
$\Delta$ vs. Qwen3-VL-32B
& $+1.5$ & $+3.3$ & $-0.1$ & $0.0$
& $+7.6$ & $+1.7$ & $+7.4$ & $+13.7$
& $+5.0$ & $+8.5$ & $+5.7$ \\
\bottomrule
\end{NiceTabular}
\end{adjustbox}
\end{table}
\FloatBarrier

Across the 4B, 8B, and 32B Qwen3-VL models, MeSD raises V-STAR mLGM
by 28.6, 24.5, and 17.2 points, respectively. The corresponding
four-benchmark averages rise by 7.2, 5.7, and 5.7 points. TVGBench
mIoU improves at every size, although the 8B model loses 1.0 point
on VideoMMMU Adaptation and the 32B model loses 0.1 point on
WorldSense Understanding. The overall gains therefore coexist with
small category-level trade-offs.

\subsection{Effects of Correction Selection.}
Table~\ref{tab:correction_selection} compares correction selection
strategies while keeping the evidence-conditioned teachers and
gated preference aggregation fixed.
Applying token-level advantage refinement to all trajectories
without FCD reduces V-STAR mLGM from 50.2 to 41.0 and
TVGBench mIoU from 29.8 to 23.0.
These results support the contribution of failure-conditioned
distillation beyond advantage refinement.
To examine the role of verification-guided selection, the
random-selection variant uniformly selects valid trajectories
with the required teacher evidence, regardless of verification
outcome.
Within each supported task and batch, the number selected
matches the number eligible for correction under its current
rollouts, while the remaining trajectories retain token-level
advantage refinement.
This comparison tests whether selecting verified failures
provides more effective corrective guidance than random selection.
\begin{table}[htbp]
\centering
\caption{
Effect of correction selection.
CR denotes credit refinement.
}
\label{tab:correction_selection}

\begin{adjustbox}{max width=\linewidth}
\begin{NiceTabular}{@{}lcccccccccc@{}}
\CodeBefore
\rowcolor{blue!10}{5}
\Body
\toprule
\multirow{2}{*}[-2.7pt]{\textbf{Variant}}
& \multicolumn{2}{c}{\textbf{Optimization Routing}}
& \multicolumn{2}{c}{\textbf{V-STAR}}
& \multicolumn{2}{c}{\textbf{WorldSense}}
& \multicolumn{2}{c}{\textbf{Video-MMMU}}
& \textbf{LRR}
& \textbf{TVGBench} \\
\cmidrule(lr){2-3}
\cmidrule(lr){4-5}
\cmidrule(lr){6-7}
\cmidrule(lr){8-9}
\cmidrule(lr){10-10}
\cmidrule(lr){11-11}
& RKL Selection
& Other Trajectories
& mAM
& mLGM
& Overall
& Understanding
& Overall
& Perception
& Acc.
& mIoU \\
\midrule

w/o FCD
& None
& CR (all)
& 30.7
& 41.0
& 38.6
& 38.7
& 53.8
& 68.3
& 71.3
& 23.0 \\

Random
& Random subset
& CR (unselected)
& 33.1
& 45.6
& 40.1
& 42.4
& 55.0
& 70.7
& 72.5
& 26.4 \\
\midrule

\textbf{MeSD}
& Verified failures (strict)
& CR (unselected)
& \textbf{35.5}
& \textbf{50.2}
& \textbf{41.6}
& \textbf{46.0}
& \textbf{56.1}
& \textbf{73.0}
& \textbf{73.6}
& \textbf{29.8} \\
\bottomrule
\end{NiceTabular}
\end{adjustbox}

\end{table}

\subsection{Training Efficiency.}
To improve training efficiency, we adopt vLLM as the inference
backend during training. As shown in
Table~\ref{tab:training_efficiency}, at the same training budget
of 2,327 steps, MeSD reduces training time from 61.36 to
20.96 hours compared with VISD no-feedback, achieving a
$2.93\times$ speedup. The corresponding compute cost decreases
from 981.70 to 335.38 A800 GPU-hours, a reduction of 65.8\%.
\begin{table}[htbp]
    \centering
    \caption{Training efficiency comparison at the same number of training steps.}
    \label{tab:training_efficiency}
    \begin{tabular}{lrrr}
        \toprule
        Method & Steps & Time (h) & A800 GPU-hours \\
        \midrule
        VISD no-feedback & 2,327 & 61.36 & 981.70 \\
        \textbf{MeSD} & 2,327 & \textbf{20.96} & \textbf{335.38} \\
        \bottomrule
    \end{tabular}
\end{table}

\subsection{Detailed Comparisons and Ablations}
\label{app:detailed_results}

Tables~\ref{tab:app_vstar_full} and~\ref{tab:app_general_full} expand
the main-text comparison into seven V-STAR metrics and the category-level
WorldSense and VideoMMMU scores, alongside LRR and TVGBench. Chain1
follows what--when--where, and Chain2 follows what--where--when.
Scores are reported to one decimal place; ``--'' denotes an unreported
result. Bold and underlining mark the best and second-best scores among
the displayed models on V-STAR and among open-source models elsewhere.

MeSD attains the highest V-STAR mAM (35.5) and mLGM (50.2), compared
with 35.1 and 48.9 for VISD. VISD remains stronger on both temporal
IoU chains (26.8 and 27.2 versus 25.8 and 25.0). On the other four
benchmarks, MeSD improves over VISD in WorldSense overall score
(41.6 versus 39.9), VideoMMMU overall score (56.1 versus 53.8),
LRR accuracy (73.6 versus 73.5), and TVGBench mIoU (29.8 versus 24.8).

\begin{table}[H]
\centering
\setlength{\belowcaptionskip}{4pt}
\caption{Detailed V-STAR comparison.}
\label{tab:app_vstar_full}
\small
\renewcommand{\arraystretch}{1.0}
\setlength{\tabcolsep}{3pt}

\begin{tabularx}{\linewidth}{@{}l*{7}{Y}@{}}
\toprule
\multirow{2}{*}{\textbf{Model}}
& \textbf{What}
& \multicolumn{2}{c}{\textbf{When (tIoU)}}
& \multicolumn{2}{c}{\textbf{Where (sIoU)}}
& \multicolumn{2}{c}{\textbf{Overall}} \\
\cmidrule(lr){3-4}
\cmidrule(lr){5-6}
\cmidrule(lr){7-8}
& Acc. & Chain1 & Chain2 & Chain1 & Chain2 & mAM & mLGM \\
\midrule
Gemini-2-Flash ~\citep{google2025gemini2flash}
& 53.0 & 24.5 & 23.8 & 4.6 & 2.2 & 26.9 & 35.6 \\
GPT-4o ~\citep{openai2024gpt4o}
& 60.8 & 16.7 & 12.8 & 6.5 & 3.0 & 26.8 & 38.2 \\
Oryx-1.5-7B ~\citep{liu2025oryx}
& 20.5 & 13.5 & 14.8 & 10.1 & 3.5 & 15.1 & 13.8 \\
InternVL-2.5-8B ~\citep{chen2024internvl25}
& 44.2 & 8.7 & 7.8 & 0.7 & 0.1 & 17.6 & 24.9 \\
Qwen2.5-VL-7B ~\citep{bai2025qwen25vl}
& 33.5 & 15.4 & 13.8 & 17.0 & 2.5 & 19.3 & 22.4 \\
TRACE ~\citep{guo2025trace}
& 17.6 & 19.1 & 17.1 & 0.0 & 0.0 & 12.0 & 13.3 \\
VideoLLaMA3-7B ~\citep{zhang2025videollama3}
& 41.9 & 23.0 & 23.1 & 0.9 & 0.2 & 21.7 & 27.0 \\
LLaVA-Video ~\citep{zhang2025llavavideo}
& 49.5 & 10.5 & 12.2 & 1.9 & 1.3 & 20.8 & 27.3 \\
VideoChat2 ~\citep{li2024mvbench}
& 36.2 & 13.7 & 12.5 & 2.5 & 1.0 & 17.0 & 20.3 \\
Open-o3-Video-7B ~\citep{meng2025openo3video}
& 61.0 & 24.5 & 24.0 & 25.4 & 6.0 & 33.7 & 46.6 \\
VISD-7B ~\citep{lin2026visd}
& \underline{61.9}
& \textbf{26.8}
& \textbf{27.2}
& \underline{26.7}
& \underline{6.3}
& \underline{35.1}
& \underline{48.9} \\
\midrule
\rowcolor{blue!10}
\textbf{MeSD-7B (Ours)}
& \textbf{63.7}
& \underline{25.8}
& \underline{25.0}
& \textbf{28.1}
& \textbf{6.5}
& \textbf{35.5}
& \textbf{50.2} \\
\bottomrule
\end{tabularx}
\end{table}
\vspace{6pt}

\begin{table}[H]
\centering
\caption{Detailed comparison on WorldSense, VideoMMMU, LRR, and TVGBench.}
\label{tab:app_general_full}
\small
\renewcommand{\arraystretch}{1.04}
\setlength{\tabcolsep}{3pt}
\begin{tabularx}{\linewidth}{@{}l*{4}{Y}@{}}
\toprule
\multirow{2}{*}{\textbf{Model}} & \multicolumn{4}{c}{\textbf{WorldSense}} \\
\cmidrule(lr){2-5}
& Overall & Recognition & Understanding & Reasoning \\
\midrule
GPT-4o & 42.6 & -- & -- & -- \\
Qwen2.5-VL-7B & 35.5 & 31.8 & 38.4 & 38.7 \\
Open-o3-Video-7B & 38.9 & 36.8 & 39.7 & \underline{40.5} \\
VISD-7B & \underline{39.9} & \textbf{39.4} & \underline{40.4} & \underline{40.5} \\
\midrule
\rowcolor{blue!10}
\textbf{MeSD-7B (Ours)} & \textbf{41.6} & \underline{38.9} & \textbf{46.0} & \textbf{42.3} \\
\bottomrule
\end{tabularx}
\par\vspace{8pt}\noindent
\begin{tabularx}{\linewidth}{@{}l>{\hsize=.55\hsize\linewidth=\hsize}Y>{\hsize=1.05\hsize\linewidth=\hsize}Y>{\hsize=1.05\hsize\linewidth=\hsize}Y>{\hsize=1.40\hsize\linewidth=\hsize}Y>{\hsize=.90\hsize\linewidth=\hsize}Y>{\hsize=1.05\hsize\linewidth=\hsize}Y@{}}
\toprule
\multirow{2}{*}{\textbf{Model}} & \multicolumn{4}{c}{\textbf{VideoMMMU}} & \textbf{LRR} & \textbf{TVGBench} \\
\cmidrule(lr){2-5}\cmidrule(lr){6-6}\cmidrule(lr){7-7}
& Overall & Perception & Adaptation & Comprehension & Accuracy & mIoU \\
\midrule
GPT-4o & 61.2 & 66.0 & -- & -- & -- & -- \\
Qwen2.5-VL-7B & 51.2 & 64.7 & 38.7 & 48.7 & 59.3 & 16.3 \\

Open-o3-Video-7B & 52.6 & 67.4 & 41.3 & 49.1 & 69.1 & 19.2 \\
VISD-7B & \underline{53.8} & \underline{69.3} & \underline{42.3} & \underline{49.7} & \underline{73.5} & \underline{24.8} \\
\midrule
\rowcolor{blue!10}
\textbf{MeSD-7B (Ours)} & \textbf{56.1} & \textbf{73.0} & \textbf{44.3} & \textbf{51.3} & \textbf{73.6} & \textbf{29.8} \\
\bottomrule
\end{tabularx}
\end{table}
\FloatBarrier

\setlength{\intextsep}{4pt plus 2pt minus 2pt}
\setlength{\textfloatsep}{6pt plus 2pt minus 2pt}

\noindent\textbf{Component and training-strategy ablations.}
\label{app:detailed_ablations}

Tables~\ref{tab:app_ablation_vstar} and~\ref{tab:app_ablation_general}
report the full component and training-strategy ablations. The variant
names follow Table~\ref{tab:ablation}: w/o FCD removes failure-conditioned
distillation; w/o ES uses one teacher with all evidence; w/o T\&S keeps
answer evidence only; w/o GPA averages teacher preferences; and w/o VGO
replaces verification-guided optimization with multi-evidence OPSD.
Bold marks the best score in each column.

Every ablated variant scores below MeSD on V-STAR mLGM and TVGBench
mIoU. Removing failure-conditioned distillation reduces these scores
from 50.2 to 41.0 and from 29.8 to 23.0, respectively. Replacing
verification-guided optimization with OPSD yields a TVGBench mIoU
of 20.7. These declines support the contribution of targeted correction
and outcome-aware training, although the w/o ES variant slightly exceeds
MeSD on VideoMMMU Adaptation (44.7 versus 44.3).

\begin{table}[H]
\centering
\setlength{\belowcaptionskip}{4pt}
\caption{Detailed V-STAR ablations.}
\label{tab:app_ablation_vstar}
\small
\renewcommand{\arraystretch}{1.0}
\setlength{\tabcolsep}{3pt}

\begin{tabularx}{\linewidth}{@{}l*{7}{Y}@{}}
\toprule
\multirow{2}{*}{\textbf{Model}}
& \textbf{What}
& \multicolumn{2}{c}{\textbf{When (tIoU)}}
& \multicolumn{2}{c}{\textbf{Where (sIoU)}}
& \multicolumn{2}{c}{\textbf{Overall}} \\
\cmidrule(lr){3-4}
\cmidrule(lr){5-6}
\cmidrule(lr){7-8}
& Acc. & Chain1 & Chain2 & Chain1 & Chain2 & mAM & mLGM \\
\midrule
\rowcolor{orange!10}
\multicolumn{8}{l}{\textit{Component Variants}} \\
w/o FCD
& 56.4 & 22.0 & 20.9 & 23.7 & 5.0 & 30.7 & 41.0 \\
w/o ES
& 56.5 & 22.1 & 21.7 & 23.4 & 5.2 & 30.9 & 41.3 \\
w/o T\&S
& 56.4 & 22.3 & 21.4 & 23.5 & 5.1 & 30.8 & 41.2 \\
w/o GPA
& 56.5 & 22.3 & 21.7 & 23.6 & 5.1 & 30.9 & 41.4 \\
\rowcolor{orange!10}
\multicolumn{8}{l}{\textit{Training Strategy Variants}} \\
Base VLM
& 33.5 & 15.4 & 13.8 & 17.0 & 2.5 & 19.3 & 22.4 \\
GSPO
& 61.0 & 24.5 & 24.0 & 25.4 & 6.0 & 33.7 & 46.6 \\
GRPO
& 55.7 & 21.9 & 21.5 & 23.6 & 5.3 & 30.6 & 40.7 \\
w/o VGO
& 56.6 & 22.4 & 19.4 & 23.2 & 5.2 & 30.6 & 40.9 \\
\midrule
\rowcolor{blue!10}
\textbf{MeSD}
& \textbf{63.7}
& \textbf{25.8}
& \textbf{25.0}
& \textbf{28.1}
& \textbf{6.5}
& \textbf{35.5}
& \textbf{50.2} \\
\bottomrule
\end{tabularx}
\end{table}
\vspace{6pt}

\begin{table}[H]
\centering
\caption{Detailed ablations on WorldSense, VideoMMMU, LRR, and TVGBench.}
\label{tab:app_ablation_general}
\fontsize{8.5}{10}\selectfont
\renewcommand{\arraystretch}{0.94}
\setlength{\tabcolsep}{3pt}
\begin{tabularx}{\linewidth}{@{}l*{4}{Y}@{}}
\toprule
\multirow{2}{*}{\textbf{Model}} & \multicolumn{4}{c}{\textbf{WorldSense}} \\
\cmidrule(lr){2-5}
& Overall & Recognition & Understanding & Reasoning \\
\midrule
\rowcolor{orange!10}
\multicolumn{5}{l}{\textit{Component Variants}} \\
w/o FCD & 38.6 & 38.2 & 38.7 & 39.0 \\
w/o ES & 38.3 & 38.4 & 38.4 & 38.1 \\
w/o T\&S & 37.9 & 37.3 & 39.3 & 37.3 \\
w/o GPA & 39.5 & 37.9 & 39.5 & 38.6 \\
\rowcolor{orange!10}
\multicolumn{5}{l}{\textit{Training Strategy Variants}} \\
Base VLM & 35.5 & 31.8 & 38.4 & 38.7 \\
GSPO & 37.5 & 36.8 & 39.7 & 40.5 \\
GRPO & 39.2 & 38.7 & 40.4 & 38.8 \\
w/o VGO & 39.5 & 38.8 & 41.0 & 38.9 \\
\midrule
\rowcolor{blue!10}
\textbf{MeSD} & \textbf{41.6} & \textbf{38.9} & \textbf{46.0} & \textbf{42.3} \\
\bottomrule
\end{tabularx}
\par\vspace{6pt}\noindent
\begin{tabularx}{\linewidth}{@{}l>{\hsize=.55\hsize\linewidth=\hsize}Y>{\hsize=1.05\hsize\linewidth=\hsize}Y>{\hsize=1.05\hsize\linewidth=\hsize}Y>{\hsize=1.40\hsize\linewidth=\hsize}Y>{\hsize=.90\hsize\linewidth=\hsize}Y>{\hsize=1.05\hsize\linewidth=\hsize}Y@{}}
\toprule
\multirow{2}{*}{\textbf{Model}} & \multicolumn{4}{c}{\textbf{VideoMMMU}} & \textbf{LRR} & \textbf{TVGBench} \\
\cmidrule(lr){2-5}\cmidrule(lr){6-6}\cmidrule(lr){7-7}
& Overall & Perception & Adaptation & Comprehension & Accuracy & mIoU \\
\midrule
\rowcolor{orange!10}
\multicolumn{7}{l}{\textit{Component Variants}} \\
w/o FCD & 53.8 & 68.3 & 43.0 & 50.0 & 71.3 & 23.0 \\
w/o ES & 55.4 & 71.0 & \textbf{44.7} & 50.7 & 70.2 & 22.4 \\
w/o T\&S & 54.8 & 71.0 & 42.0 & \textbf{51.3} & 70.6 & 22.3 \\
w/o GPA & 53.9 & 69.7 & 43.0 & 49.0 & 70.4 & 22.9 \\
\rowcolor{orange!10}
\multicolumn{7}{l}{\textit{Training Strategy Variants}} \\
Base VLM & 51.2 & 64.7 & 38.7 & 48.7 & 59.3 & 16.3 \\
GSPO & 52.3 & 68.0 & 41.3 & 49.1 & 69.4 & 20.8 \\
GRPO & 53.2 & 69.3 & 40.0 & 50.3 & 71.9 & 22.5 \\
w/o VGO & 54.1 & 67.7 & 44.3 & 50.3 & 71.2 & 20.7 \\
\midrule
\rowcolor{blue!10}
\textbf{MeSD} & \textbf{56.1} & \textbf{73.0} & 44.3 & \textbf{51.3} & \textbf{73.6} & \textbf{29.8} \\
\bottomrule
\end{tabularx}
\end{table}

\subsection{Training Curves}
\label{app:training_curves}
\setlength{\abovecaptionskip}{2pt}
\setlength{\intextsep}{2pt}
\captionsetup[figure]{skip=2pt}

Reward rises quickly and then levels off at all three Qwen3-VL sizes
(Figure~\ref{fig:app_qwen3vl_reward}). MeSD also maintains higher
reward than the w/o ES, w/o T\&S, and w/o GPA variants
(Figure~\ref{fig:app_component_reward}), consistent with the
ablation results in Tables~\ref{tab:app_ablation_vstar}
and~\ref{tab:app_ablation_general}.

\begin{figure}[H]
    \centering
    \includegraphics[width=0.85\textwidth]{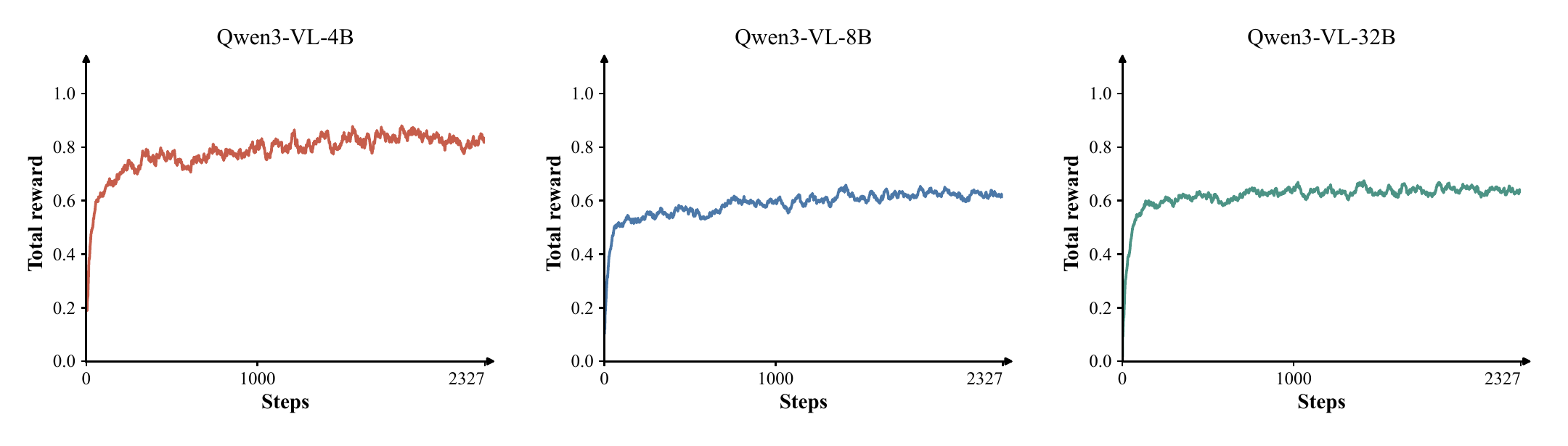}
    \caption{Training reward for Qwen3-VL models at 4B, 8B, and 32B.}
    \label{fig:app_qwen3vl_reward}
\end{figure}

\begin{figure}[H]
    \centering
    \includegraphics[width=0.85\textwidth]{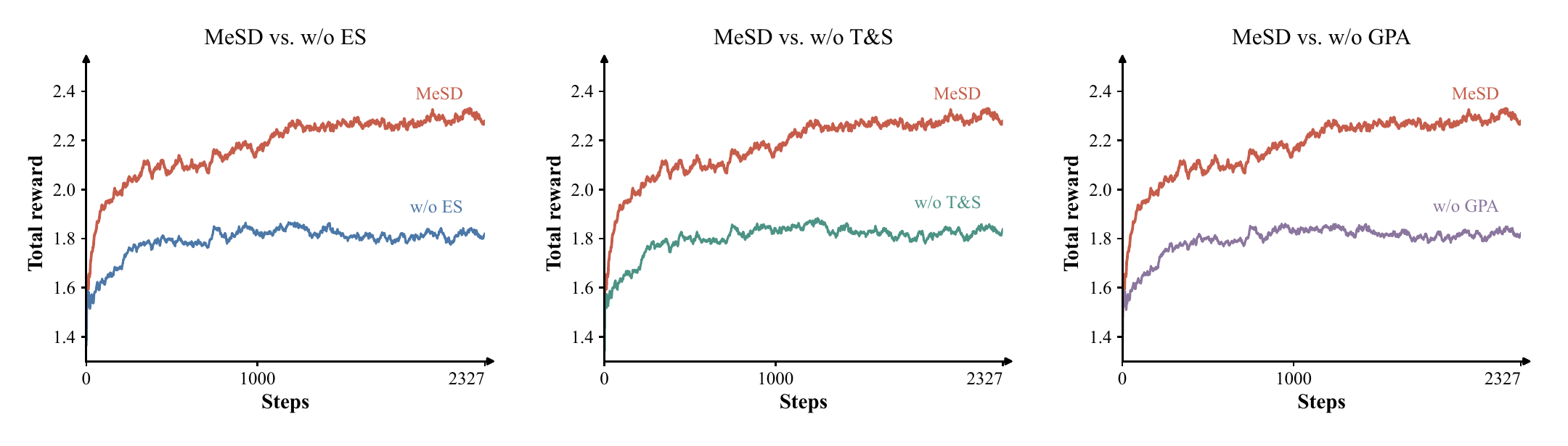}
    \caption{Training reward for MeSD and the w/o ES, w/o T\&S, and
    w/o GPA component variants.}
    \label{fig:app_component_reward}
\end{figure}

\FloatBarrier
\endgroup
\clearpage

\section{Case Study}
\noindent\begin{minipage}{\linewidth}
\noindent\textbf{Object--action grounding.}
Figure~\ref{fig:case_baby} shows MeSD identifying the baby as the actor and the toy as the object being bitten.
The generated reasoning describes the action at multiple timestamps, and the final answer is consistent with the ground truth.
Separate references to the baby and the toy link the textual description to model-predicted regions, allowing readers to inspect their correspondence in the sampled frames.
\par\smallskip
{\centering
\captionsetup{font=small,skip=3pt}
\includegraphics[width=0.8\linewidth]{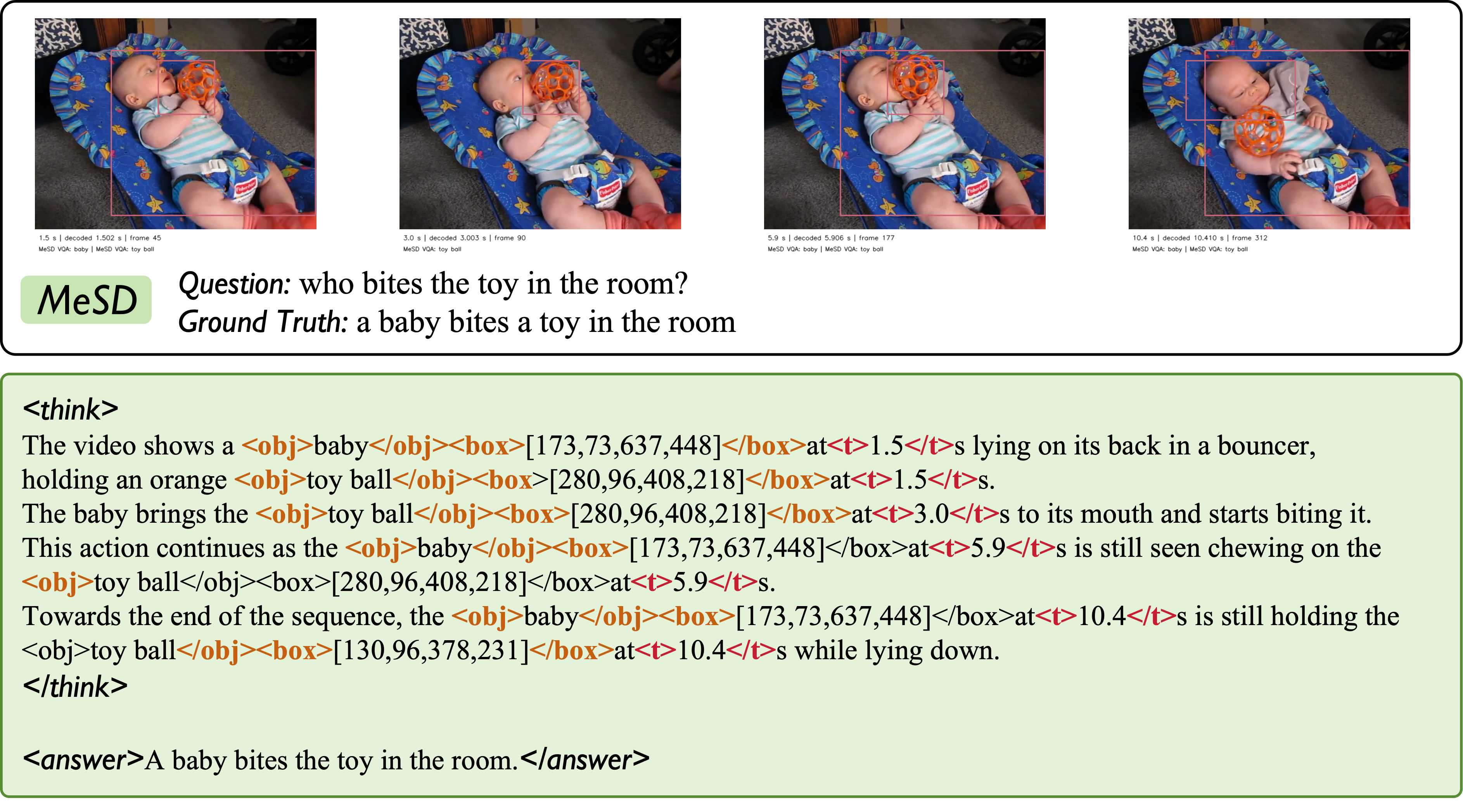}
\captionof{figure}{Object--action grounding. MeSD identifies the baby biting the toy and includes model-predicted boxes and timestamps in its reasoning.}
\label{fig:case_baby}
\par}
\end{minipage}
\par\vspace{10pt}
\noindent\begin{minipage}{\linewidth}
\noindent\textbf{Person--interaction grounding.}
Figure~\ref{fig:case_hug} shows MeSD distinguishing the man from the child and describing how he holds the child at 2.4\,s and 9.8\,s.
The final answer identifies the adult in white and is semantically consistent with the ground truth.
Explicit references to both participants clarify their respective roles and provide spatial and temporal cues for assessing the described interaction.
\par\smallskip
{\centering
\captionsetup{font=small,skip=3pt}
\includegraphics[width=0.8\linewidth]{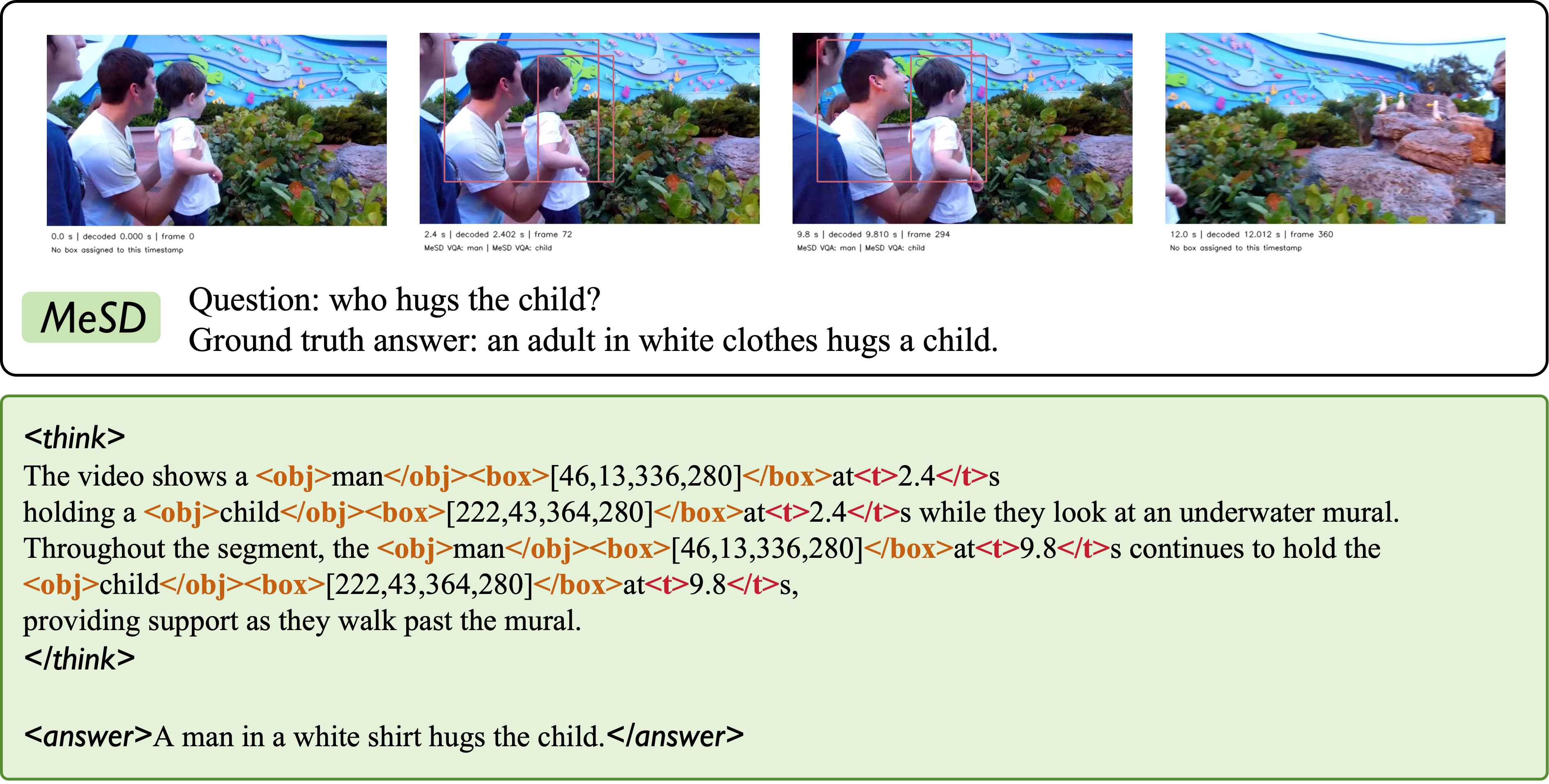}
\captionof{figure}{Person--interaction grounding. MeSD identifies the man holding the child, with predicted regions corresponding to the two participants referenced in its reasoning.}
\label{fig:case_hug}
\par}
\end{minipage}
\clearpage
\noindent\begin{minipage}{\linewidth}
\noindent\textbf{Answer correctness and temporal grounding.}
Figure~\ref{fig:case_failure} shows that MeSD produces an answer accepted by the evaluator for the identity question, while temporal localization remains less precise.
Both reasoning chains predict 19--21\,s, compared with the annotated interval of 0--5.6\,s, yielding zero temporal IoU.
This contrast highlights the complementary value of evaluating answer correctness and temporal grounding separately.
MeSD's explicit temporal predictions make this discrepancy directly inspectable, providing a concrete target for further refinement.
\par\smallskip
{\centering
\captionsetup{font=small,skip=3pt}
\includegraphics[width=0.8\linewidth]{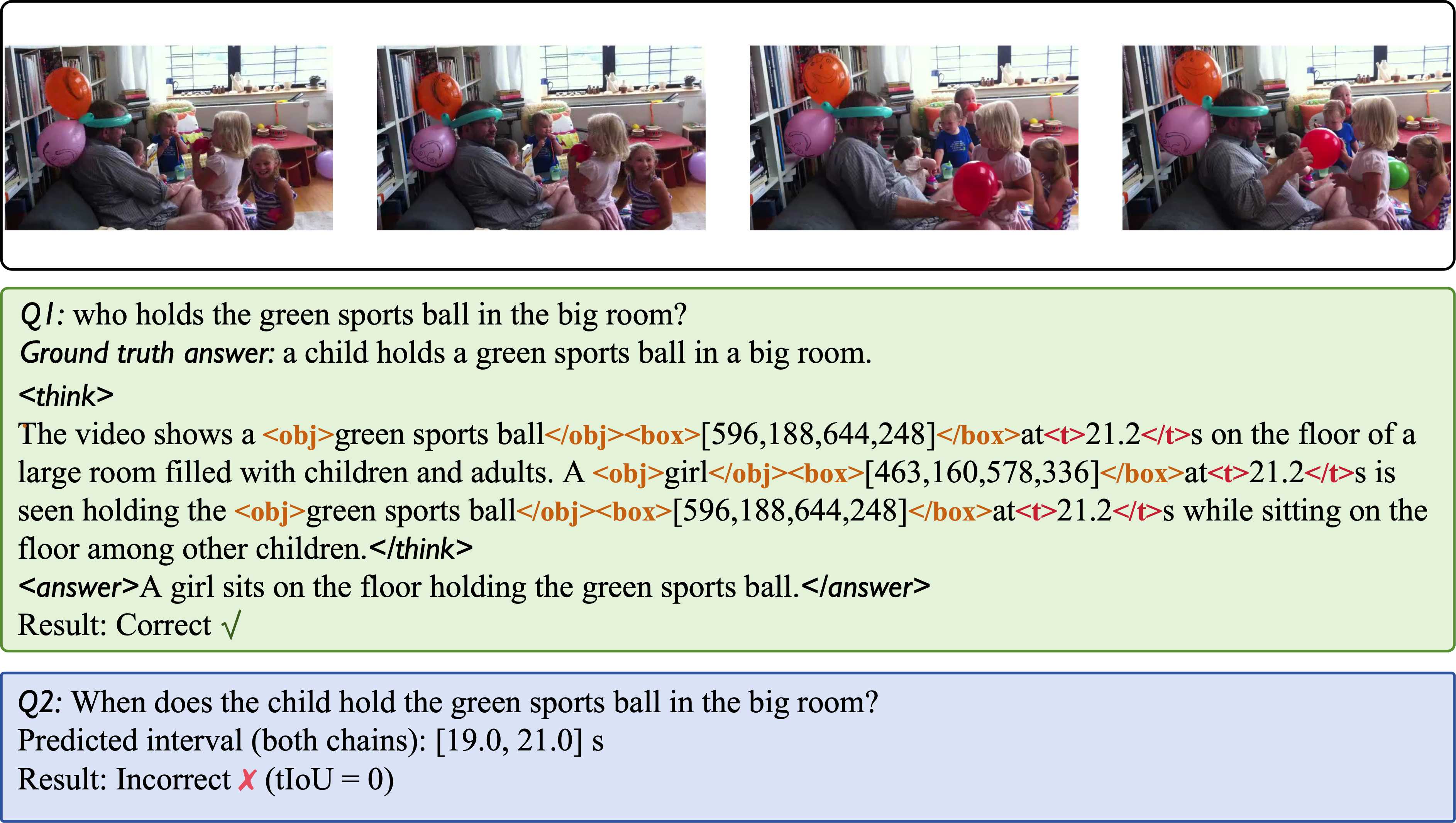}
    \captionof{figure}{Temporal grounding failure. MeSD's answer to the identity question is judged correct, but both reasoning chains predict a temporal interval that does not overlap with the ground truth.}
    \label{fig:case_failure}
\par}
\end{minipage}
\par\medskip



\section{Limitations and Future Work}
\label{app:limitations}
While MeSD effectively integrates answer, temporal, and spatial annotations, the current study explores only a subset of the evidence that could support video reasoning. Its multi-evidence formulation offers a natural basis for incorporating richer supervision, but the potential benefits of additional annotation types remain unexplored. The present evaluation therefore leaves open how greater annotation diversity and finer granularity could further improve self-distillation.

Future work could introduce complementary annotations, such as object trajectories and event relations, to construct additional evidence-conditioned teacher views. Extending the fusion mechanism to account for differences in evidence granularity and reliability could help integrate these sources effectively. Systematically examining the contributions of different annotation types and their combinations would further clarify how richer supervision supports video reasoning. Such extensions may provide more detailed guidance for complex temporal and spatial reasoning and broaden the applicability of MeSD.

\stopcontents[appendix]

\end{document}